\documentclass[10pt,twocolumn,letterpaper]{article}

\usepackage[pagenumbers]{cvpr} 

\usepackage{graphicx}
\usepackage{amsmath}
\usepackage{amssymb}
\usepackage{booktabs}

\usepackage[accsupp]{axessibility}
\usepackage{listings}
\usepackage[frozencache]{minted}
\usepackage{xcolor}
\usepackage{textcomp}
\usepackage{mathtools}
\usepackage{multirow}
\usepackage{array}
\usepackage{empheq}
\usepackage[autostyle]{csquotes}
\usepackage{nicefrac}
\usepackage{bbm}
\usepackage{amsfonts}
\usepackage{epsdice}
\usepackage{nth}
\usepackage{float}

\setminted[python]{ %
    linenos=true,             
    autogobble=true,          
    frame=lines,
    framesep=2mm,
    fontsize=\footnotesize,
    escapeinside=||
}

\definecolor{green}{HTML}{00aa00}
\definecolor{blue}{HTML}{0000aa}
\definecolor{red}{HTML}{aa0000}
\usepackage[pagebackref,breaklinks,colorlinks,citecolor=green,linkcolor=red,urlcolor=blue]{hyperref}

\usepackage[capitalize]{cleveref}
\crefname{section}{Sec.}{Secs.}
\Crefname{section}{Section}{Sections}
\Crefname{table}{Table}{Tables}
\crefname{table}{Tab.}{Tabs.}

\def\cvprPaperID{1048} 
\def\confName{CVPR}
\def\confYear{2022}

\newcommand\vcdice[1]{\vcenter{\hbox{\epsdice{#1}}}}
\let\op\operatorname

\newcommand{\eat}[1]{}

\newcommand{\wo}{\mathit{w\!/\!o}}

\newcommand{\norm}[1]{\left\lVert#1\right\rVert}
\DeclareMathOperator*{\argmax}{arg\,max}

\DeclarePairedDelimiter\parentheses{\lparen}{\rparen}

\DeclareMathOperator*{\Categorical}{Categorical}

\DeclareMathOperator{\Softmax}{Softmax}

\DeclarePairedDelimiterX\set[1]\lbrace\rbrace{#1}

\newcommand{\Cat}[2]{\Categorical \parentheses*{#1,\,#2}}

\newcommand{\smallsim}{\smallsym{\mathrel}{\sim}}

\newcommand{\vect}[1]{\boldsymbol{#1}}

\makeatletter
\newcommand{\smallsym}[2]{#1{\mathpalette\make@small@sym{#2}}}
\newcommand{\make@small@sym}[2]{%
  \vcenter{\hbox{$\m@th\downgrade@style#1#2$}}%
}
\newcommand{\downgrade@style}[1]{%
  \ifx#1\displaystyle\scriptstyle\else
    \ifx#1\textstyle\scriptstyle\else
      \scriptscriptstyle
  \fi\fi
}

\DeclareRobustCommand\onedot{\futurelet\@let@token\@onedot}
\def\@onedot{\ifx\@let@token.\else.\null\fi\xspace}

\newenvironment{code}
 {\RecustomVerbatimEnvironment{Verbatim}{BVerbatim}{}%
  \def\FV@BProcessLine##1{%
    \hbox{%
      \hbox to\z@{\hss\theFancyVerbLine\kern\FV@NumberSep}%
      \FancyVerbFormatLine{##1}%
    }%
  }%
  \VerbatimEnvironment
  \setbox\z@=\hbox\bgroup
  \begin{minted}{python}}
 {\end{minted}\egroup
  \leavevmode\vbox{\hrule\kern2mm\box\z@\kern2mm\hrule}}

\makeatother

\begin{document}

\title{Unified Multivariate Gaussian Mixture for Efficient Neural Image Compression}

\author{Xiaosu Zhu\textsuperscript{1}
\and
Jingkuan Song\textsuperscript{1}\thanks{Corresponding author.}
\and
Lianli Gao\textsuperscript{1}
\and
Feng Zheng\textsuperscript{2}
\and
Heng Tao Shen\textsuperscript{1}
\and
\textsuperscript{1}Center for Future Media, University of Electronic Science and Technology of China \\
\textsuperscript{2}Southern University of Science and Technology \\
{\tt\small
\href{mailto:xiaosu.zhu@outlook.com}{xiaosu.zhu@outlook.com}, \href{mailto:jingkuan.song@gmail.com}{jingkuan.song@gmail.com}, \href{mailto:shenhengtao@hotmail.com}{shenhengtao@hotmail.com}
}
}
\maketitle

\begin{abstract}
Modeling latent variables with priors and hyperpriors is an essential problem in variational image compression. Formally, trade-off between rate and distortion is handled well if priors and hyperpriors precisely describe latent variables. Current practices only adopt univariate priors and process each variable individually. However, we find  inter-correlations and intra-correlations exist when observing latent variables in a vectorized perspective. These findings reveal visual redundancies to improve rate-distortion performance and parallel processing ability to speed up compression. This encourages us to propose a novel vectorized prior. Specifically, a multivariate Gaussian mixture is proposed with means and covariances to be estimated. Then, a novel probabilistic vector quantization is utilized to effectively approximate means, and remaining covariances are further induced to a unified mixture and solved by cascaded estimation without context models involved. Furthermore, codebooks involved in quantization are extended to multi-codebooks for complexity reduction, which formulates an efficient compression procedure. Extensive experiments on benchmark datasets against state-of-the-art indicate our model has better rate-distortion performance and an impressive $3.18\times$ compression speed up, giving us the ability to perform real-time, high-quality variational image compression in practice. Our source code is publicly available at \url{https://github.com/xiaosu-zhu/McQuic}.
\end{abstract}

\section{Introduction}
\label{Sec.Intro}
As a crucial technique in image processing, lossy image compression has been studied for an extended period~\cite{Survey1,Survey2,Survey3,Survey4}. The goal is to achieve high perceptual reconstruction performance, extreme compression rate, and efficient processing pipeline. Classical lossy image compression standards, \eg, JPEG~\cite{JPEG,JPEG2000}, BPG~\cite{BPG}, HEIF~\cite{HEVC}, VVC~\cite{VVC}, have been widely applied and adopted as fundamental components in almost all image processing software. However, the explosion of multimedia content in the digital era still raises urgent requests to find an effective and efficient compressor to tackle storage costs.

\begin{figure}[t]
  \centering
  \begin{subfigure}[b]{0.45\linewidth}
  \includegraphics[width=\linewidth]{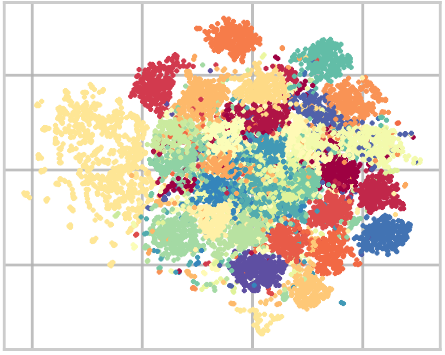}
  \caption{$\smash{\vect{y}^1}$}
  \label{Fig.UMAPVisualization-Left}
  \end{subfigure}
  \hfil
  \begin{subfigure}[b]{0.45\linewidth}
  \includegraphics[width=\linewidth]{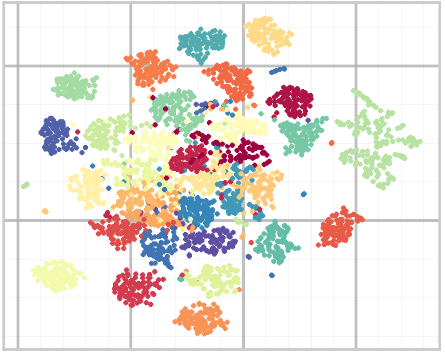}
  \caption{$\smash{\vect{y}^2 = \left(\vect{y}^1 - \vect{\mathfrak{y}}^1\right)_\downarrow}$}
  \label{Fig.UMAPVisualization-Right}
  \end{subfigure}
  \caption{UMAP~\cite{UMAP} projection of $128$-d latent vectors with a toy $2$-level $32$-codeword model from $24$ Kodak images.
  \textbf{Left}: Latent vectors extracted from analysis transform are correlated and can be described by multivariate Gaussian mixture. \textbf{Right}: Next level's latents are under similar distribution.}
  \label{Fig.UMAPVisualization}
\end{figure}

\begin{figure*}
  \centering
  \begin{subfigure}[b]{0.15\linewidth}
  \includegraphics[width=\linewidth]{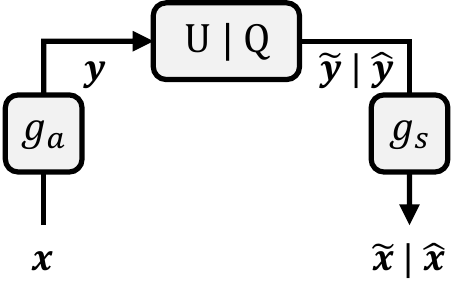}
    \caption{Factorized prior~\cite{FactorizedPrior}.}
    \label{Fig.Diagram1}
  \end{subfigure}
  \hfill
  \begin{subfigure}[b]{0.1677\linewidth}
  \includegraphics[width=\linewidth]{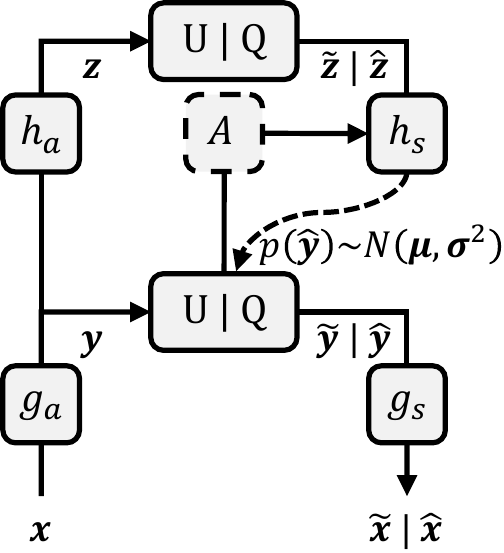}
    \caption{Hyperprior~\cite{ScaleHyper,JointHyper,GlobalReference}.}
    \label{Fig.Diagram2}
  \end{subfigure}
  \hfill
  \begin{subfigure}[b]{0.23225\linewidth}
  \includegraphics[width=\linewidth]{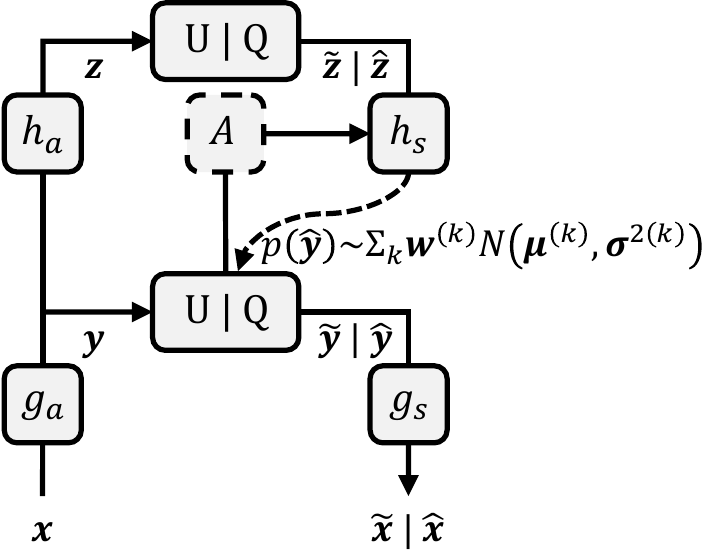}
    \caption{Discretized Gaussians~\cite{GMMAttention}.~~~~~~~~~~~~}
    \label{Fig.Diagram3}
  \end{subfigure}
  \hfill
  \begin{subfigure}[b]{0.26416\linewidth}
  \includegraphics[width=\linewidth]{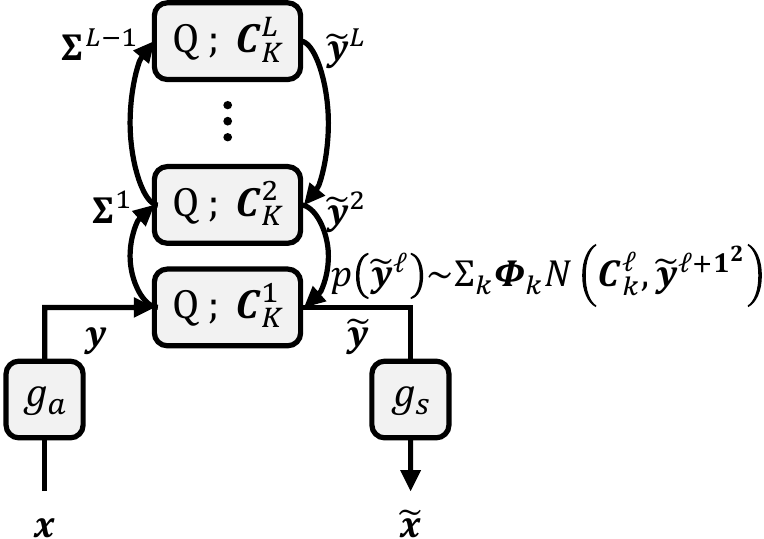}
    \caption{Vectorized prior.~~~~~~~~~~~~~~~~~~~~~~~~~~~}
    \label{Fig.Diagram4}
  \end{subfigure}
  \hfill
  \caption{Operational diagrams of different methods. We generalize prior as a unified multivariate Gaussian mixture.}
  \label{fig:short}
\end{figure*}

Distinct from the above traditional codecs, learnable neural image compression is proposed by exploiting advantages of deep neural networks. It adopts neural networks as nonlinear transforms to extract binaries from images and restore them, while essential research problem is to handle the trade-off between rate and distortion~\cite{RDTheory}. Recent studies propose variational image compression and arrange above trade-off as a Lagrange multiplier for joint optimization~\cite{FactorizedPrior,ScaleHyper,Conditional,JointHyper,GMMAttention}. They introduce univariate priors and hyperpriors to describe latent variables and make a breakthrough to control rate. We summarize advances in this task as a series of operational diagrams in \crefrange{Fig.Diagram1}{Fig.Diagram3}.

To design an effective compressor in variational image compression, an appropriate prior that precisely describes quantized latent variables is needed~\cite{ScaleHyper,JointHyper,GMMAttention}. \cref{Fig.UMAPVisualization-Left} demonstrates observation of latent variables grouped by channels. This vectorized perspective reveals correlations of latents that help us to find a prior. Note that latent vector comes from a specific region of an image and represents this region's visual appearance, correlations between vectors can be summarized as inter-correlation and intra-correlation. Inter-correlation comes from facts that images have spatial redundancy~\cite{GlobalReference} \ie vectors extracted from visually-similar regions or patches are closed together. Meanwhile, similar regions still have differences in details, resulting in intra-correlation \ie covariances. Two properties guide us to find a vectorized prior which could describe two correlations by means and covariances.

Univariate priors previous works adopt may not be sufficient to describe above observations, because they process each scalar value individually and lack a whole view over vectors. In other words, adopting a vectorized prior mainly has two impacts. Firstly, it treats latents as vectors along channels other than scalars, helping to summarize inter- and intra-correlations. Secondly, vectorized processing has the potential to speed up compression procedure. Therefore in this paper, we propose a novel vectorized prior for variational image compression. Specifically, a unified multivariate Gaussian mixture is proposed to describe latents. Then, a probabilistic vector quantization with cascaded estimation is designed to effectively and efficiently estimate means and covariances without context models involved. Multi-codebooks are further incorporated into quantization to reduce complexity and enable flexible rate control. The whole procedure is demonstrated in \cref{Fig.Diagram4} and our contribution is summarized below:

1. We propose a new vectorized perspective for variational image compression. Unlike previous works, ours considers correlations between latent vectors and formulates a unified multivariate Gaussian mixture. We further propose a probabilistic vector quantization with cascaded estimation to estimate means and covariances.

2. A multi-codebook structure is further incorporated into quantization to reduce complexity and enable flexible rate control. Overall framework is able to perform effective and efficient compression with the help of vectorized prior.

3. Extensive experiments on benchmark datasets reveal impacts of vectorized prior. Compared to state-of-the-art, our method achieves better rate-distortion performance with an impressive $3.18\times$ speed up for compression latency. These results reveal possibility to provide practical variational image compression with vectorized prior.

\begin{figure*}[t]
  \centering
  \begin{subfigure}[b]{0.47\linewidth}
  \includegraphics[width=\linewidth]{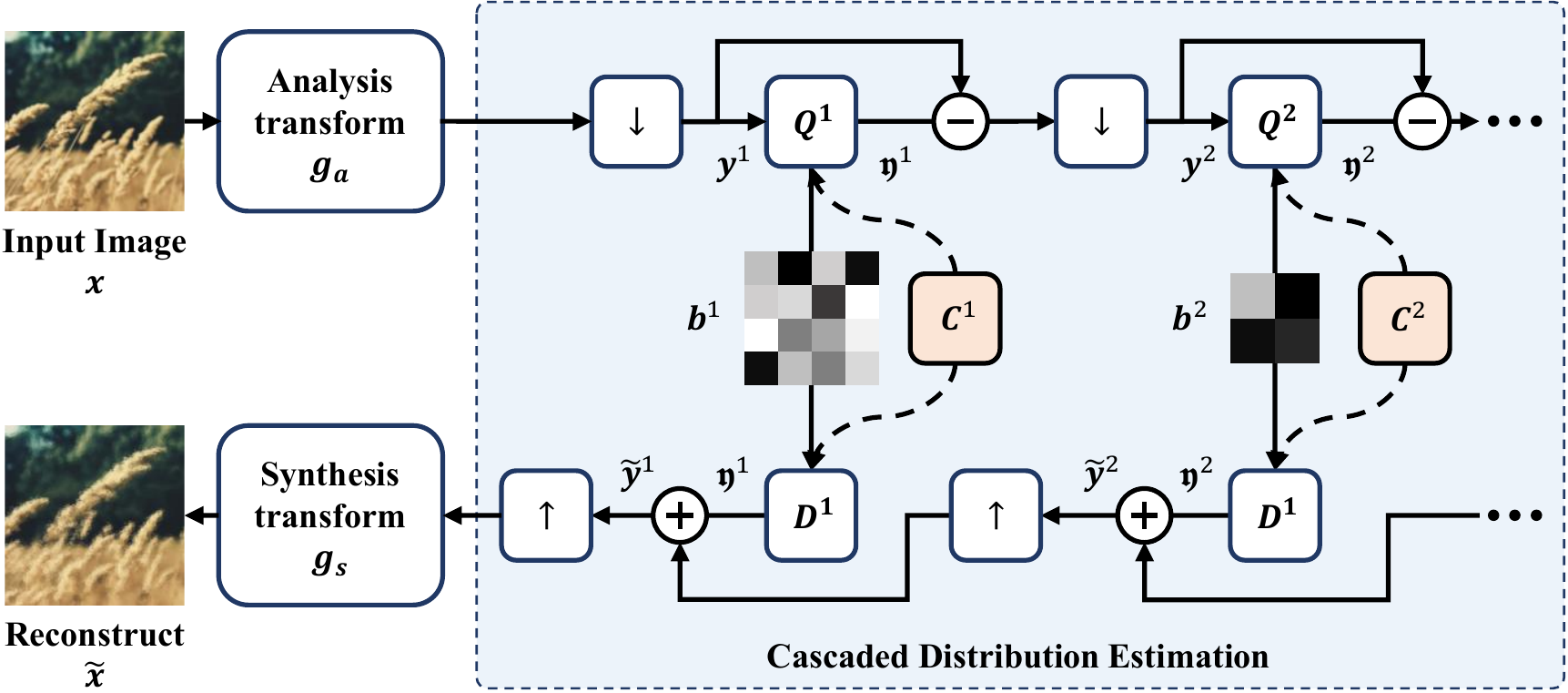}
  \caption{}
  \label{Fig.Framework}
  \end{subfigure}
  \hfil
  \begin{subfigure}[b]{0.38\linewidth}
  \includegraphics[width=\linewidth]{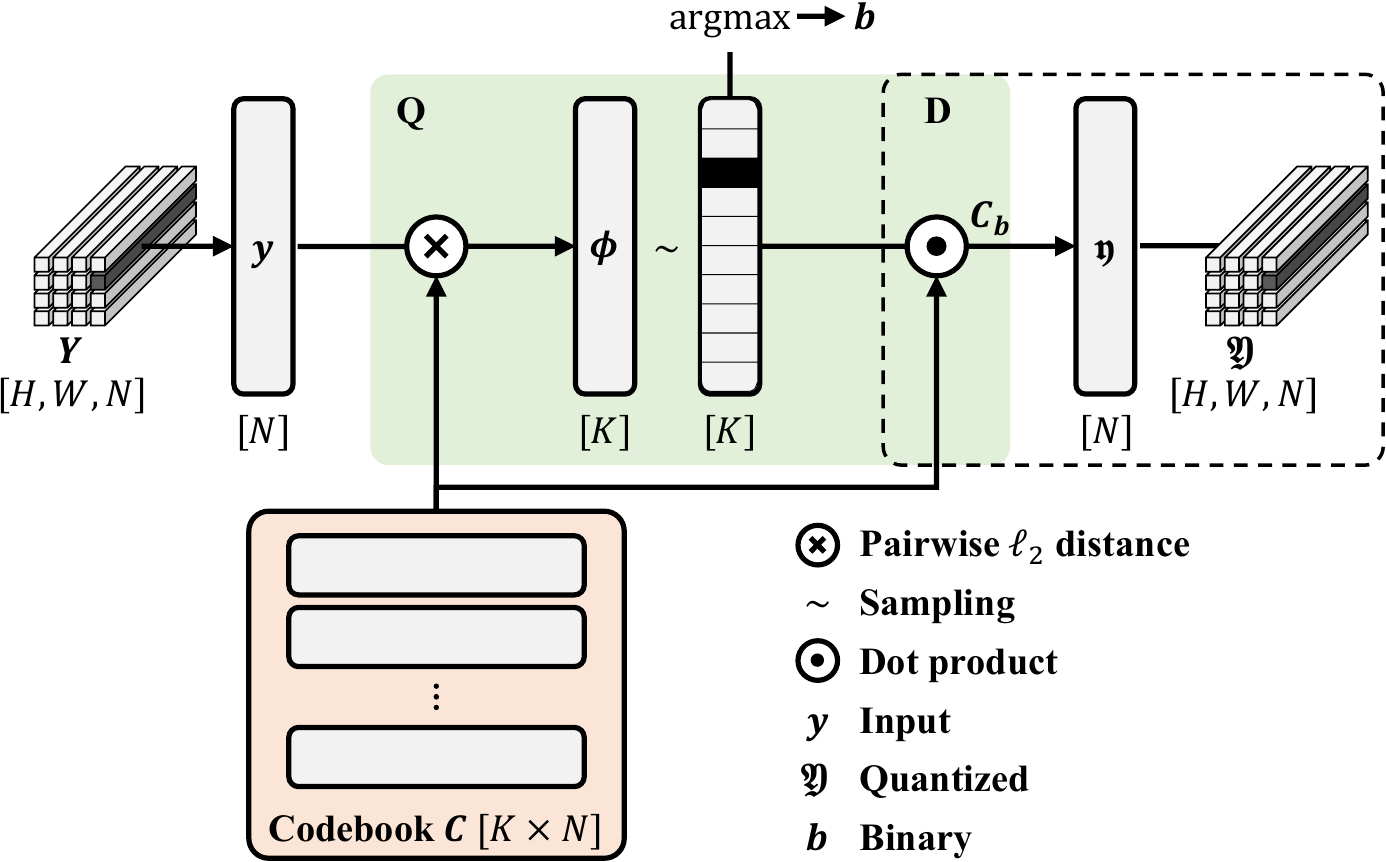}
  \caption{}
  \label{Fig.Quantizer}
  \end{subfigure}
  \caption{(a) Proposed network utilizes cascaded estimation with probabilistic vector quantization ($\mathit{Q}$) and reverse ($\mathit{D}$) to model vectorized prior. \enquote{$\downarrow$}, \enquote{$\uparrow$} denotes down- and up-sampling blocks. (b) Proposed probabilistic vector quantization constructs Categorical distribution parameterized by $\vect{\phi}$ to sample $\vect{b}$ and quantize $\vect{y}$.}
\end{figure*}

\section{Related Works}
This paper focuses on variational image compression. Formally, this approach utilizes an auto-encoder to process latents in order to compress images. Studies focus on handling trade-off between rate and distortion. Specifically, latents are quantized by rounding to the nearest integer~\cite{FactorizedPrior} or prototype~\cite{Conditional} in order to perform entropy coding with \eg range coder.  Ball{\'e}~\etal~\cite{FactorizedPrior} propose an entropy model and train network end-to-end (\cref{Fig.Diagram1}). Subsequently, Hyperprior model~\cite{ScaleHyper} performs variational inference by hyperprior prediction. \cref{Fig.Diagram2,Fig.Diagram3} give two mainstream styles of hyperpriors. The first~\cite{ScaleHyper,JointHyper} is under a shifted and scaled Gaussian distribution, while the second~\cite{GMMAttention} generalizes distribution to Gaussian mixture. Both of them could take an auxiliary context model~\cite{JointHyper,GMMAttention,GlobalReference,PixelCNN} for precise estimation and further reduce compression rate.

Other than scalar quantization they adopt, a vector quantization~(VQ) is adapted to our proposed vectorized prior. Studies on VQ for image compression have a long history early to 1980s~\cite{VQCompress1,VQCompress2,VQCompress3}. The core problem of VQ to integrate into deep networks is to tackle the non-differentiable $\argmax$ operation involved in quantization. Agustsson~\etal~\cite{SoftQuantization} relax $\argmax$ to $\Softmax$ and propose a soft-to-hard end-to-end quantization. Van den Oord~\etal~\cite{VQVAE} and Esser~\etal~\cite{VQGAN} instead utilize a straight-through estimator and directly pass quantized latents to decoder. Similar approaches are also applied to many other tasks~\cite{DPgQ,UBGAN,beit}.

\section{Proposed Method}

In this section, we firstly give preliminaries and overall demonstration of our proposed method.

Given an arbitrary image $\vect{x}$, variational image compression takes an analysis transform $g_{a}$ to produce latent variable $\vect{y} = \op{g_{a}}\left(\vect{x}\right)$, which will be quantized $\hat{\vect{y}} = \op{q}\left(\vect{y}\right)$. A synthesis transform $g_{s}$ restores $\hat{\vect{x}} = \op{g_{s}}\left(\hat{\vect{y}}\right)$ from $\hat{\vect{y}}$. Distortion between $\vect{x}$ and $\hat{\vect{x}}$ is measured by a perceptual metric $\op{d}\left(\vect{x}, \hat{\vect{x}}\right)$. Meanwhile, size of compressed $\hat{\vect{y}}$ is controlled by an entropy model $p_{\hat{\vect{y}}}$. Therefore, trade-off between rate:
\begin{align}
\label{Eq.Rate}
    \min_{g_{a}, g_{s}}{\mathcal{R}} &= \mathbb{E}_{\vect{x}} \left[ -\log_2{ p_{\hat{\vect{y}}}\left(\hat{\vect{y}}\right) } \right] \\
\intertext{and distortion:}
\label{Eq.Distortion}
    \min_{g_{a}, g_{s}}{\mathcal{D}} &= \mathbb{E}_{\vect{x}} \left[ \op{d}\left(\vect{x}, \hat{\vect{x}}\right) \right]
\end{align}
is the essential optimization objective. To enable end-to-end training, above compressed values $\hat{\cdot}$ are approximated by $\tilde{\cdot}$.

\textbf{Framework.} We put overall framework in \cref{Fig.Framework}. Specifically, analysis and synthesis transforms are similar as \cite{GMMAttention} where residual and attention blocks are involved. Then, to quantize and transmit latent variables, cascaded estimation adopts a series of down-sampling or up-sampling blocks followed by probabilistic vector quantization $\mathit{Q}$ or dequantization $\mathit{D}$, respectively. Take a look at a single $Q^\ell$ at level $\ell$, it accepts latent $\vect{y}^\ell \subseteq \mathbb{R}^{h^\ell \times w^\ell \times N}$ with $N$ channels, $h^\ell \times w^\ell$ size, then produces intermediate latent $\vect{\mathfrak{y}}^\ell$ in same shape using codebook $\vect{C}^\ell \subseteq \mathbb{R}^{K \times N}$. Corresponding binary code $\vect{b}^\ell \subseteq \set{0, 1}^{h^\ell \times w^\ell \times \log_2{K}}$ is transmitted to the decoder side, and residual $\vect{y}^\ell - \vect{\mathfrak{y}}^\ell$ is passed to the next level.

$D^\ell$ does symmetrical thing. It restores $\vect{\mathfrak{y}}^\ell$ by $\vect{C}^\ell_{\vect{b}^\ell}$. Then, $\vect{\mathfrak{y}}^\ell$ and upper level $\tilde{\vect{y}}^{\ell + 1}$ are added up to get $\tilde{\vect{y}}^\ell$. Therefore, the core pipeline of encoding and decoding is defined as following recursive functions:
\begin{align}
\label{Eq.Recursive1}
    &\left\{
    \begin{alignedat}{-1}
        \left(\vect{\mathfrak{y}}^\ell,\;\vect{b}^\ell\right)&=\op{Q^\ell}\left( \vect{y}^\ell;\;\vect{C}^\ell\right), \\
        \vect{y}^{\ell+1}&={\left(\vect{y}^\ell - \vect{\mathfrak{y}}^\ell\right)}_{\downarrow},\;1 \leq \ell \leq L,
    \end{alignedat}
    \right.\\
\label{Eq.Recursive2}
    &\left\{
    \begin{alignedat}{-1}
\;\;\;\;\;\;\;\;\;\;\vect{\mathfrak{y}}^\ell   &=\op{D^\ell}\left( \vect{b}^\ell;\;\vect{C}^\ell \right), \\
\;\;\;\;\;\;\;\;\;\;\tilde{\vect{y}}^\ell &=\vect{\mathfrak{y}}^\ell + \left(\tilde{\vect{y}}^{\ell+1}\right)_{\uparrow} , \; 1 \leq \ell < L,
    \end{alignedat}
    \right.
\end{align}
where ${\left(\cdot\right)}_{\downarrow}, {\left(\cdot\right)}_{\uparrow}$ denote down-sampling and up-sampling.

Explaining these equations requires us to give definition of vectorized prior~(\cref{Sec.VectorizedPrior}), way to perform quantization and estimation~(\cref{Sec.LSQ}) and a generalization on prior by cascaded estimation~(\cref{Sec.RecursiveEstimators}).

\subsection{Unified Multivariate Gaussian Mixture}
\label{Sec.RecursiveGMM}
\subsubsection{Vectorized Prior}
\label{Sec.VectorizedPrior}

An intuition to work with $\vect{y}^1$ is to group it by channels: $\vect{Y} = \set{\vect{y}^1_j \subseteq \mathbb{R}^{N}, 1 \leq j \leq h^{\ell}w^{\ell}}$ where $j$ is the spatial location in latent feature map. For simplicity, we rearrange $\vect{y} \in \vect{Y}$ as a $N$-dim vector. Such arrangement helps to define $p_{\vect{Y}}\left(\vect{y}\right)$ as a mixture of $N$-dim multivariate Gaussians:
\begin{equation}
\label{Eq.GMM}
\begin{split}
    p_{\vect{Y}}\left(\vect{y}\right) &= \sum_{k=1}^{K}{\vect{\Phi}_{k} \mathcal{N}\left(\vect{\mu}_k, \vect{\Sigma}_k\right)}, \\
    \mathit{where}\;\vect{\Phi} &\sim \Cat{K}{\vect{\phi}}.
\end{split}
\end{equation}
Here, $\vect{\mu}_k$ and $\vect{\Sigma}_k$ are mean and covariance matrix of the $k$-th Gaussian component. $\vect{\Phi}$ represents a mixture parameterized by $K$-Categorical distribution with un-normalized log-probabilities ${\vect{\phi}}$.

The given vectorized prior is based on two kinds of correlations we summarize from $\vect{y}$. \cref{Fig.UMAPVisualization-Left} reveals these by UMAP projection with $\vect{y}$ that directly extracted from backbone. Firstly, inter-correlations between $\vect{y}$s show similarities or visual redundancies \ie extracted latent vectors are close if their original visual pattern are similar. This helps to cluster $\vect{y}$s into several distinct Gaussian components where cluster centroids are equivalent to means $\vect{\mu}_k$. Secondly, vectors clustered in a same component are not identical but have covariance $\vect{\Sigma}_k$, since they still have subtle differences. To further quantize vectors in $\vect{Y}$, a vector quantization to estimate $\vect{\mu}$ and $\vect{\Sigma}$ is needed.

\subsubsection{Probabilistic Vector Quantization}
\label{Sec.LSQ}

We propose a learnable, probabilistic vector quantization that makes an \textit{approximation} on above distribution, which is demonstrated in \cref{Fig.Quantizer}. Specifically, it maintains a codebook $\vect{C} \subseteq \mathbb{R}^{K \times N}$ consists of $K$ codewords. Input $\vect{y}$ is quantized by assigning a specific codeword to it, which is expressed as the following discrete conditional distribution:
\begin{equation}
\label{Eq.Phi}
\begin{split}
    p_{\vect{\mathfrak{Y}} \mid \vect{Y}}{\left( \vect{\mathfrak{y}} \mid \vect{y};\;\vect{C} \right)} &= \prod_{k=1}^{K}{\zeta{\left(\vect{\phi}\right)}_k^{\mathbbm{1}\left\{\vect{\mathfrak{y}} = \vect{C}_k\right\}}}, \\
    \mathit{where}\;\vect{\phi}_{k} &= -\norm{\vect{y} - \vect{C}_k}_2^2,\;1 \leq k \leq K.
\end{split}
\end{equation}
Correspondingly, $\vect{\mathfrak{Y}}$ is set of centroids $\vect{\mathfrak{y}}$. $p_{\vect{\mathfrak{Y}} \mid \vect{Y}}$ formulates a Categorical distribution where $\vect{\mathfrak{y}}$ is assigned to the $k$-th codeword with probability $\zeta{\left(\vect{\phi}\right)}_k$. $\zeta$ the $\Softmax$ function, $\vect{\phi}$ the negative Euclidean distance between $\vect{y}$ and codeword, $\mathbbm{1}\left\{ \cdot \right\}$ the characteristic function. To obtain $\vect{\mathfrak{y}}$, we sample above distribution:
\begin{equation}
    \vect{\mathfrak{y}} \sim \op{Q}\left(\vect{y};\;\vect{C}\right) = p_{\vect{\mathfrak{Y}} \mid \vect{Y}}{\left( \vect{\mathfrak{y}} \mid \vect{y};\;\vect{C} \right)}
\end{equation}
which results in one-of-$K$ codeword of $\vect{C}$. Intuitively, probability to choose $\vect{C}_k$ will be high if $\vect{y}$ is close to $\vect{C}_k$.

After a sample is drawn from $p_{\vect{\mathfrak{Y}} \mid \vect{Y}}$, $\vect{b}$ is immediately obtained by index of picked codeword, which will be encoded into binary stream to transmit. On decoding side, $\mathit{D}$ retrieves identical picked codeword by $\vect{C}_{\vect{b}}$ to restore $\vect{\mathfrak{y}}$ since codebook $\vect{C}$ is a shared parameter between $\mathit{Q}$ and $\mathit{D}$.

Above quantization defines a probabilistic model. By minimizing \cref{Eq.Distortion}, codewords in $\vect{C}$ is derived to approximately estimate means of Gaussian components of $p_{\vect{Y}}$\footnote{The proof is placed in the supplementary materials.}:
\begin{equation}
\label{Eq.MeanEstimation}
    \vect{C}_k \colonapprox \mathbb{E}\set{\vect{y} \in \vect{Y} \mid \vect{\Phi}_k = 1} = \vect{\mu}_k
\end{equation}
which automatically perform alignment between codewords and means. Compared to commonly used $k$-means, the proposed quantization chooses codeword stochastically other than directly pick the nearest one in a deterministic way. It models partial of $p_{\vect{Y}}(\vect{y})$ and aggregates into codebook. Moreover, introduced randomness may help network to escape the local optima during training.

\subsubsection{Cascaded Estimation}
\label{Sec.RecursiveEstimators}

It is worth noting that above proposed quantization is unable to estimate covariance matrix $\vect{\Sigma}$ according to previous derivation. Noticed that:
\begin{equation}
\label{Eq.Covariance}
\begin{split}
    \vect{\Sigma}_k &= \mathbb{E}\left\{\left(\vect{Y}_k - \vect{\mu}_k\right)\left(\vect{Y}_k - \vect{\mu}_k\right)^\intercal\right\}, \\
    \mathit{where}\;\vect{Y}_k &= \set{\vect{y} \in \vect{Y} \mid \vect{\Phi}_k = 1}.
\end{split}
\end{equation}

An intuition is raised to tackle this by designing a residual connection, since:
\begin{equation}
\label{Eq.Residual}
\begin{split}
    \mathbb{E}\left\{\vect{y} - \vect{\vect{\mathfrak{y}}} \mid \vect{\Phi}_k = 1 \right\} &= \mathbb{E}\left\{ \vect{Y}_k - \vect{C}_k \right\}\\
    &\approx \mathbb{E}\left\{ \vect{Y}_k - \vect{\mu}_k \right\}.
\end{split}
\end{equation}
That is why \cref{Eq.Recursive1,Eq.Recursive2} are proposed. We take former level's $\vect{y} - \vect{\mathfrak{y}}$ as inputs of latter level, and let latter level's neural network to predict $\vect{\Sigma}$. \cref{Fig.UMAPVisualization-Right} tells us a trick to assume residuals on every level to be also under Gaussian mixture, helping us to expand \cref{Eq.GMM} and give completed definition of the \textbf{\textit{unified multivariate Gaussian mixture}}:
\begin{align}
\label{Eq.Unified2}
    p_{\vect{Y}^{\ell} \mid \vect{Y}^{\ell + 1}}\left(\vect{y}^{\ell} \mid \vect{y}^{\ell + 1}\right) &= \sum_{k=1}^{K}{\vect{\Phi}^{\ell}_{k} \mathcal{N}\left(\vect{\mu}^{\ell}_k, \vect{y}^{\ell + 1}\right)}\\
\intertext{and model the compressed signal $\tilde{\vect{y}}$ by:}
    p_{\tilde{\vect{Y}}^{\ell} \mid \tilde{\vect{Y}}^{\ell + 1}}\left(\tilde{\vect{y}}^{\ell} \mid \tilde{\vect{y}}^{\ell + 1}\right) &= \sum_{k=1}^{K}{\vect{\Phi}^{\ell}_{k} \mathcal{N}\left(\vect{C}^{\ell}_k, \tilde{\vect{y}}^{\ell + 1}\right)}.
\end{align}
We should emphasize that \enquote{$\vect{y}^{\ell + 1}$}, \enquote{$\tilde{\vect{y}}^{\ell + 1}$} here are not strictly covariance matrices but are used to estimate covariance. Restoration of $\tilde{\vect{y}}$ starts from $\tilde{\vect{y}}^L$, and produces $\tilde{\vect{y}}^\ell$ level-by-level according to \cref{Eq.Recursive2}.

\begin{figure}[t]
  \centering
  \includegraphics[width=0.8\columnwidth]{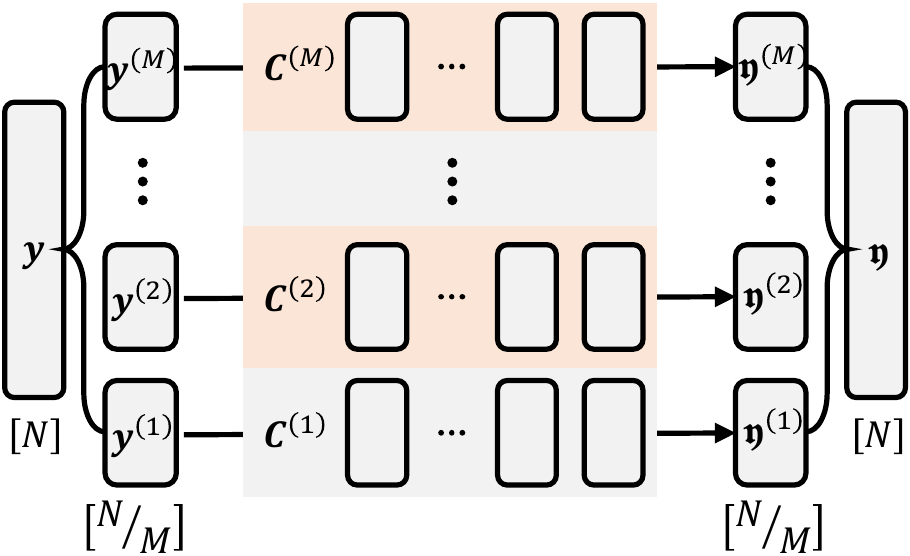}
  \caption{Multi-codebook structure. $\vect{y}$ is split into $M$ groups and quantize them separately with sub-codebooks. Each sub-codebook parameterizes an individual distribution to model $\vect{y}^{\left(m\right)}$.}
  \label{Fig.MCQuantizer}
\end{figure}

\subsection{Reduce Complexity with Multi-Codebooks}
\label{Sec.MCQ}
We could handle above quantization by maintaining a codebook $\vect{C}^\ell$ on each level. If all of them have codebook size $K$, codebook size will be $L \cdot K \cdot N$ and output $\vect{b}$ for any vector has a maximum bit-length of $\log_2{K}$. Unfortunately, $K$ is not allowed to be extremely large otherwise network is unaffordable heavy. Considering trade-off between model complexity and compress ability, we further utilize multi-codebooks to generalize our method. As \cref{Fig.MCQuantizer} shows, $\vect{y}^{\ell}$ is sliced into $M$ groups along channels. Each piece $\vect{y}^{\left(\ell, m\right)}$ is quantized by individual sub-codebook $\vect{C}^{\left(\ell, m\right)}$ whose total size is still $L \cdot K \cdot M \cdot \nicefrac{N}{M} = L \cdot K \cdot N$.

Introduced multi-codebook structure has several impacts. Firstly, since each part $\vect{y}^{\left(\ell, m\right)}$ has a choice out of $K$ codewords, the set of all possible combinations of codebook $\vect{C}^{\left(\ell\right)}$ is a Cartesian product of sub-codebooks:
\begin{equation}
    \vect{C}^{\left(\ell\right)} = \vect{C}^{\left(\ell, 1\right)} \times \vect{C}^{\left(\ell, 2\right)} \times \cdots \times \vect{C}^{\left(\ell, M\right)}
\end{equation}
which makes the maximum bit length become $M\log_2{K} = \log_2{K^M}$ with significantly small size of codebook $M \cdot K$. Secondly, with multi-codebooks, we could generalize \cref{Eq.Unified2} to be a combination of several individual multivariate Gaussian mixture. $M = 1$ gives the original \cref{Eq.Unified2}, while $M = N$ degenerates to univariate prior.

$L, K, M$ are hyper-parameters for us to control rate. In practice, introducing multi-codebooks will not significantly downgrade performance with much smaller codebook size compared to $M = 1$ quantization under same bit-length.

\subsection{Compression}
\label{Sec.EncodingAndDecoding}
At inference time, encoding and decoding are composed as follows: On encoder side, latents are quantized and binaries are rolled out by greedy assignments:
\begin{align}
\label{Eq.Lookup}
    \vect{b}^{\left(\ell, m\right)} &= \argmax_k{-\norm{\vect{y}^{\left(\ell, m\right)} - \vect{C}_k^{\left(\ell, m\right)}}_2}, \\
    \vect{\mathfrak{y}}^{\left(\ell, m\right)} &= \vect{C}^{\left(\ell, m\right)}_{\vect{b}^{\left(\ell, m\right)}},
\end{align}
which consumes $\mathcal{O}\left(K \cdot \nicefrac{N}{M}\right)$ time complexity for encoding a single vector. $\vect{b}^{\left(\ell, m\right)}$ is compressed based on estimated occurrence frequency. As for decoder, restoration of $\vect{\mathfrak{y}}^{\left(\ell, m\right)}$ only involves $\mathcal{O}\left(1\right)$ lookup according to \cref{Eq.Lookup}. Last but not least, these operations are highly paralleled which is GPU-friendly, gives us ability to perform high-efficient encoding and decoding in actual developments.

\section{Discussions}
In this section, we handle a few questions about model design and compare our proposed method with other works.

\textbf{Training.} The model is trained in an end-to-end manner. However, to achieve this, our quantization (\cref{Sec.LSQ}) utilizes stochastic computation graph for sampling, which is intractable to optimize. Fortunately, there are many studies to handle it. In our experiments, Gumbel reparameterization with straight-through estimator~\cite{GumbelSoftmax} has the best performance. Overall optimization is formulated as follows:
\begin{equation}
  \mathcal{L} = \mathcal{D} = \op{d}\left(\vect{x}, \tilde{\vect{x}}\right),\; \Theta \leftarrow \Theta - \eta\nabla_\Theta{\mathcal{L}},
\end{equation}
where $\Theta$ is the set of all trainable parameters in network and $\eta$ is learning-rate. Such optimization can be done by any gradient-based optimizers.

\begin{table}[t]
\centering
\resizebox{0.8\linewidth}{!}{
\begin{tabular}{@{}c|c|c|c|c|c@{}}\toprule
\textnumero & $N$                    & $L$                  & $M$  & $K$                                      & $\sup\mathit{bpp}$ \\\midrule\midrule
1           & \multirow{2}{*}{$128$} & \multirow{5}{*}{$3$} & $2$  & \multirow{5}{*}{$\left[8192, 2048, 512\right]$} & $0.1274$           \\
2           &                        &                      & $6$  &                                        & $0.3823$           \\ \cmidrule(){2-2}
3           & \multirow{3}{*}{$192$} &                      & $9$  &                                        & $0.5098$           \\
4           &                        &                      & $12$ &                                        & $0.7646$           \\
5           &                        &                      & $16$ &                                        & $1.0195$          \\\bottomrule
\end{tabular}
}
\caption{Model specifications target different rates. Empirically, we set $N=128$ for small models while $192$ for large. $L=3$ and $K=\left[8192, 2048, 512\right]$ for all models achieves expected results with affordable model sizes. $M$ is varied from $2$ to $16$ to control $\mathit{bpp}$. Theoretical upper bounds of $\mathit{bpp}$ are in the last column.}
\label{Tab.Setup}
\end{table}

\textbf{Controlling the size of compressed binaries.} The above objective only involves distortion but not rate. The reason is based on how we control size of compressed binaries, which is determined by $\vect{b}$. As aforementioned, the theoretical upper bound size of $\vect{b}$ is derived as $\sum_{l}{M \cdot \log_{2}{K} \cdot h^\ell \cdot w^\ell}$ for all levels and all groups. Different from previous works, this upper bound is much smaller (which will be revealed in \cref{Sec.Experiments}). We benefits from this to control bit rate by varying $L$, $M$, $K$ or adjusting latent feature map size $h^\ell, w^\ell$. Then the rate of encoded binaries will gradually approach theoretical upper bound as training progresses without explicit objective to control it.

\textbf{Relations to hyperprior models.} Proposed method has a strong relation to hyperprior models. Minnen~\etal~\cite{JointHyper} and Cheng~\etal~\cite{GMMAttention} also model quantized latents as a Gaussian mixture while our approach extends it to $N$-dim multivariate. If we set $M = N$, then our prior is degenerated to univariate version. The key differences is: Firstly, our vectorized prior provide rich statistics by $\vect{\mu}$ and $\vect{\Sigma}$ to describe latents and summarize visual redundancies. Secondly, side information $\vect{\mu}$ and $\vect{\Sigma}$ are automatically estimates by probabilistic vector quantization and cascaded estimation. In practice, they are sufficient to perform decoding without context model involved to give a speed up for compression.

\begin{figure*}[t]
  \centering
  \includegraphics[width=0.95\linewidth]{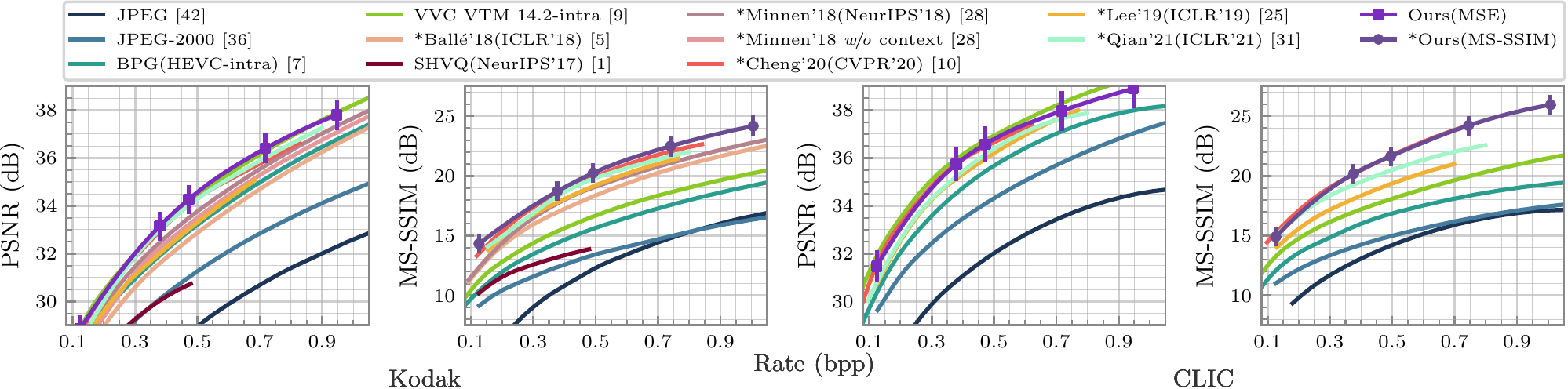}
  \caption{R-D curves on Kodak (left 2) and CLIC valid set (right 2). *: Models are optimized for MS-SSIM when with MS-SSIM metric.}
  \label{Fig.RdCurve}
\end{figure*}

\textbf{Relations to other VQ-based generative models.} There are a few works on compressing or generating images with help of VQ, \eg, SHVQ~\cite{SoftQuantization}, VQ-VAE(-2)~\cite{VQVAE,VQVAE2} and VQ-GAN~\cite{VQGAN}. Generally, they employ a $k$-means style quantizer which assigns the closest codeword to latent as we have discussed in \cref{Sec.LSQ}. In order to perform end-to-end training, codebook is updated by two-stage E-M style algorithms or straight-through estimators. Nevertheless, ours includes covariance of latents while theirs could not handle. Furthermore, our framework generalizes quantization by multi-codebook structure other than a global codebook.

Proposed multi-codebook structure shares similar ideas with product quantization~\cite{PQ}, group convolution~\cite{GroupConv} and multi-head attention~\cite{SelfAttention}. They are widely applied to vision/language tasks for rich feature learning with low costs.

\section{Experiments}
\label{Sec.Experiments}
We conduct extensive experiments to evaluate effectiveness and efficiency of our proposed method. Specifically, we first show R-D performance comparisons with other methods. Then, we measure encoder and decoder latency to demonstrate the network efficiency. Other analysis \ie ablation study and visualization are further given.

\subsection{Setup}
\textbf{Training datasets.} The training dataset is a chosen subset of \textbf{ImageNet}~\cite{ImageNet} combined with \textbf{CLIC}~\cite{CLIC} Professional training set. Specifically, we filter images from ImageNet to have more than one million pixels and randomly sample $7,415$ images from them. The whole CLIC training set with $585$ images is merged ($8,000$ images in total).

\textbf{Model Specs.} Our method to be tested is model \textnumero $1\smallsim5$ targeting different rate by varying codebook sizes. The choices consider model complexity by tuning $N$, $M$, $K$ and $L$, placed in \cref{Tab.Setup}. To train the model, we adopt LAMB optimizer~\cite{Lamb}. Training images are random-cropped to $512\times512$ and batched into $8$. Initial learning rate is set at $2 \times 10^{-3}$ and annealed to $2 \times 10^{-6}$ at end with cosine learning rate scheduler for $1,000$ epochs. All experiments are conducted with a single NVIDIA V100 GPU. The model is implemented with PyTorch~\cite{PyTorch}.

\begin{table}[t]
\centering
\resizebox{\columnwidth}{!}{
\begin{tabular}{@{}cc|rrrr@{}}
\toprule
\multicolumn{2}{c|}{\multirow{3}{*}{Methods}}              & \multicolumn{4}{c}{Latency ($\mathit{ms}$)}               \\
\multicolumn{2}{c|}{}                                      & \multicolumn{2}{c}{Encoder} & \multicolumn{2}{c}{Decoder} \\ \cmidrule(lr){3-4}\cmidrule(lr){5-6}
\multicolumn{2}{c|}{}                                      &\multicolumn{1}{c}{Abs}&\multicolumn{1}{c}{Rel}&\multicolumn{1}{c}{Abs}&\multicolumn{1}{c}{Rel}\\ \midrule\midrule
\multicolumn{2}{c|}{Ball{\'e}'18}                   & $30.66$  & $1.09\times$& $35.54$  & $1.21\times$ \\\midrule
\multirow{3}{*}{Minnen'18}&$\wo$                    & $32.89$  &$1.17\times$ & $36.24$  &$1.24\times$ \\
                               &$\rightarrow$       & $2656.66$ &$94.58\times$& $1799.47$&$61.36\times$ \\
                             &$\vcdice{5}$          & $59.13$  &$2.11\times$ & $40.40$  &$1.38\times$  \\\midrule
\multirow{2}{*}{Cheng'20}&$\rightarrow$             & $2697.58$&$96.04\times$& $1835.80$&$62.60\times$ \\
                           &$\vcdice{5}$            & $94.11$  &$3.35\times$ & $88.04$  &$3.00\times$  \\\midrule
\multicolumn{2}{c|}{Ours}                           & $\mathbf{28.09}$  &$\mathbf{1.00\times}$ & $\mathbf{29.32}$  &$\mathbf{1.00\times}$  \\\bottomrule
\end{tabular}
}
\caption{Encoding and decoding latency comparisons for image size $768 \times 512$. For theirs, we test context-free~(the firs two row) and context-enabled~(row $3\smallsim6$) models. \enquote{$\rightarrow$} means serial context model~\cite{JointHyper} while \enquote{$\vcdice{5}$} denotes parallel~\cite{Checkerboard}. Our model is \textnumero~5. Ours is the fastest model, with up to $79.32\times$ and $3.18\times$ speed up than two kinds of context-enabled models for whole compression, respectively. Ours is even faster than context-free models, since they need more than one passes to encode and decode latents.}
\label{Tab.Time}
\end{table}

\begin{figure*}[t]
  \centering
  \includegraphics[width=0.7\linewidth]{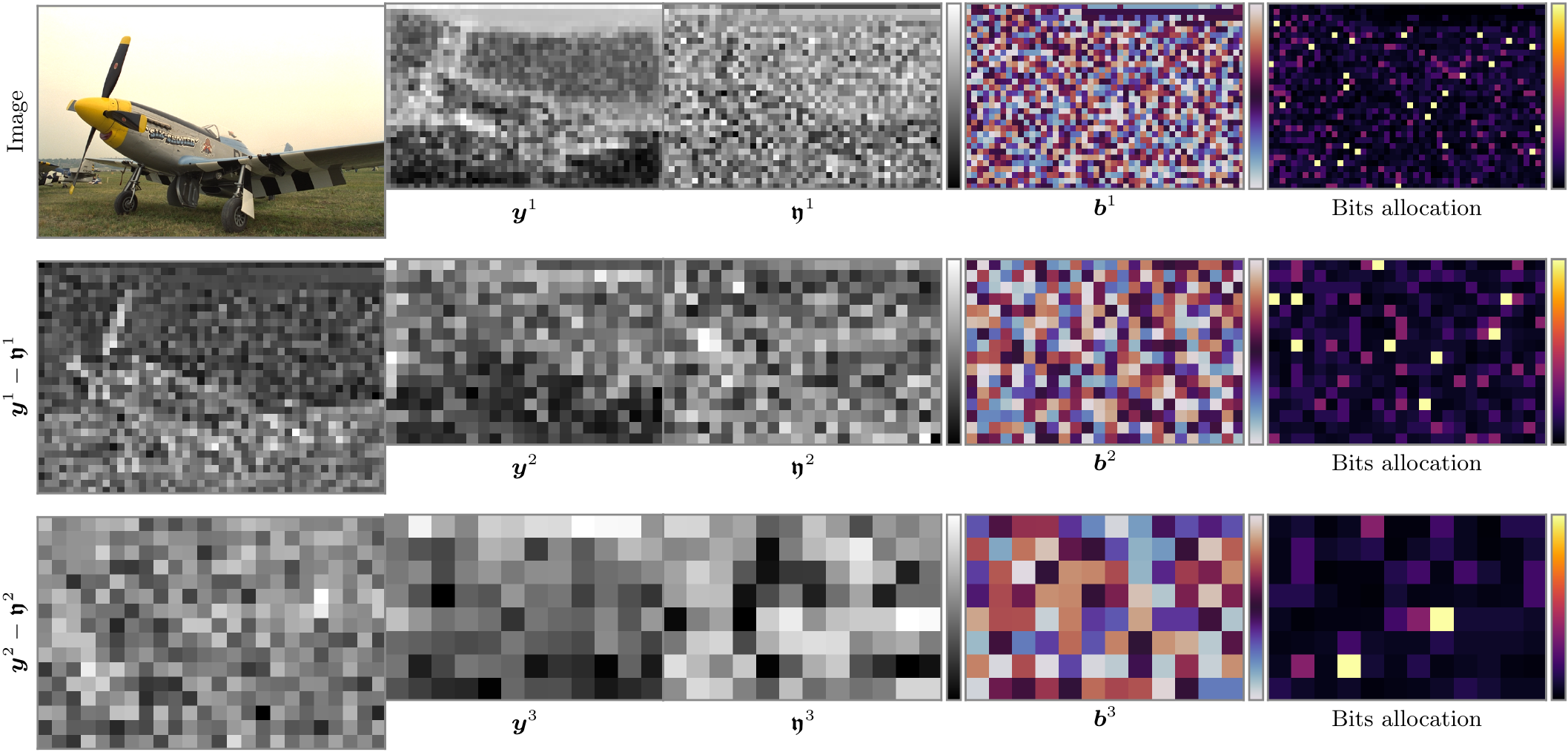}
  \caption{Visualization for a $3$-level model. $\vect{y}$ is extracted latent, $\vect{\mathfrak{y}}$ is quantized latents. By calculating $\left(\vect{y}^\ell - \vect{\mathfrak{y}}^\ell\right)$, visual redundancies are removed. $\vect{b}$ is corresponding binary (index of picked codewords). Brighter pixels in the last column mean more bits allocation.}
  \label{Fig.Latent}
\end{figure*}

\subsection{Rate-Distortion Performance}
\label{Sec.RD}

To show rate-distortion performance, rate-distortion~(R-D) points are observed and R-D curves are plotted. Specifically, to measure rate, bits-per-pixel~($\mathit{bpp}$) is calculated\footnote{They use various ways to control it, resulting in various $\mathit{bpp}$.}. While for distortion, we adopt two perceptual metrics: \textbf{PSNR} and \textbf{MS-SSIM}~(converted to decibels by $-10 \cdot \log_{10}\left(1 - \mathit{value}\right)$). Tests involve two image sets: \textbf{Kodak}~\cite{Kodak}~($24$ images) and \textbf{CLIC} Professional valid set~($41$ images). Methods to compare include a few famous traditional standards: JPEG~\cite{JPEG}, JPEG~2000~\cite{JPEG2000}, BPG~\cite{BPG}, an upcoming new standard: VVC~VTM~14.2~\cite{VVC} and $6$ deep image compression models: SHVQ~\cite{SoftQuantization}, Ball{\'e}'18~\cite{ScaleHyper}, Minnen'18~\cite{JointHyper}, Lee'19~\cite{ContextAdaptive}, Qian'21~\cite{GlobalReference} and Cheng'20~\cite{GMMAttention}. The R-D points are obtained from either public benchmarks or their paper\footnote{\url{https://github.com/tensorflow/compression}. If not specified, models are trained using corresponding distortion metrics.}. For Minnen'18, both context-free and context-involved results are reported. Since two datasets have a few images, we adopt jackknife resampling and estimation strategy to report mean value and standard error on the plot by error bars~\cite{jackknife}. More comparisons are provided in supplementary materials.

Results of Kodak and CLIC are shown in \cref{Fig.RdCurve}, respectively. For Kodak, ours outperforms state-of-the-art, while for CLIC, our model has nearly the same performance with the best deep method. Specifically, since ours adopts the same backbone as Cheng'20's, the key component to affect R-D performance is our proposed quantizers and cascaded estimators. From results we confirm our components do not hinder performance and show same or even better compress ability with state-of-the-art. Furthermore, ours achieve state-of-the-art performance without context model involved. This not only indicates effectiveness of the vectorized prior but also removes a bottleneck that slow down compression, which will be revealed in next section. Also, as rates increase, our model has a steady performance. This indicates introduced multi-codebooks are able to scale to large models by increasing $M$ to provide satisfying performance with affordable codebook size.

\subsection{Encoding and Decoding Latency}
\label{Sec.Latency}
Evaluating encoding and decoding latency reveals model efficiency, which is important in actual developments. To conduct such test, following models are adopted: Ball{\'e}'18, Minnen'18 (\enquote{$\wo$}, \enquote{$\rightarrow$}, \enquote{$\vcdice{5}$}), and Cheng'20 (\enquote{$\rightarrow$}, \enquote{$\vcdice{5}$}). Specifically, \enquote{$\wo$} means no context model involved, and \enquote{$\rightarrow$}, \enquote{$\vcdice{5}$} are serial~\cite{PixelCNN} and parallel~\cite{Checkerboard} context model variants. Our model to be tested is \textnumero~$5$. To precisely measure the latency, we feed a batch of images from Kodak with size $768 \times 512$ and track the CUDA events of encoder and decoder separately. Measurements are based on their public models or reimplementations\footnote{Tested latencies of \cite{Checkerboard} are slightly slower than their report.}.

As \cref{Tab.Time} shows, our network is the fastest method among all other models. In particular, compared to models utilizing context, ours achieves up to $79.32\times$ faster than the serials and $3.18\times$ faster than the parallels for whole compression, respectively. This efficiency gap comes from our introduced cascaded estimation that do not need context model. Furthermore, our model is even faster than context-free models \ie Ball{\'e}'18 and Minnen'18~($\mathit{\wo}$), based on how we perform (de)quantization. Ours only involves $\mathcal{O}\left(K \cdot \nicefrac{N}{M}\right)$ to quantize and $\mathcal{O}\left(1\right)$ to dequantize, and is highly paralleled running in GPU. Meanwhile, our encoders and decoders only require one forward pass, but they need two or more. In summary of \cref{Sec.RD,Sec.Latency}, our model achieves better R-D performance with an impressive compression latency, enabling us the ability to perform practical image compression with our vectorized prior.

\begin{table}[t]
\centering
\begin{tabular}{@{}r|r|cc@{}}\toprule
\multirow{2}{*}{Variants}&\multirow{2}{*}{BD-rate}&\multicolumn{2}{c}{Latency}                     \\\cmidrule(lr){3-4}
                         &                      &Encoder                 &Decoder                  \\\midrule\midrule
$\wo$ cascaded          & $8.87\%$             & $27.13$                & $28.29$                 \\
$2$-levels               & $2.33\%$             & $27.62$                & $28.77$                 \\
$4$-levels               & $-0.64\%$            & $28.93$                & $30.27$                 \\\cmidrule{3-4}
$\mathit{one}$-codebook  & $24.40\%$            &\multirow{5}{*}{$28.09$}&\multirow{5}{*}{$29.32$} \\
$\mathit{cos}$-quantizer & $4.64\%$             &                        &                         \\
\cite{VQGAN}-quantizer   & $16.20\%$            &                        &                         \\
\cite{SoftQuantization}-quantizer   & $11.48\%$          &                        &                         \\
Ours                     &     -                &                        &                         \\\bottomrule
\end{tabular}
\caption{Ablation study on $6$ variants where BD-rate \wrt original model (lower is better) and latency is reported. The first three row vary level $L$, \enquote{$\mathit{one}$-codebook} uses a global codebook. And the \nth{5}$\smallsim$\nth{7} rows modify quantizers' formulation.}
\label{Tab.AblationResult}
\end{table}

\begin{figure*}[t]
  \centering
  \includegraphics[width=0.71\linewidth]{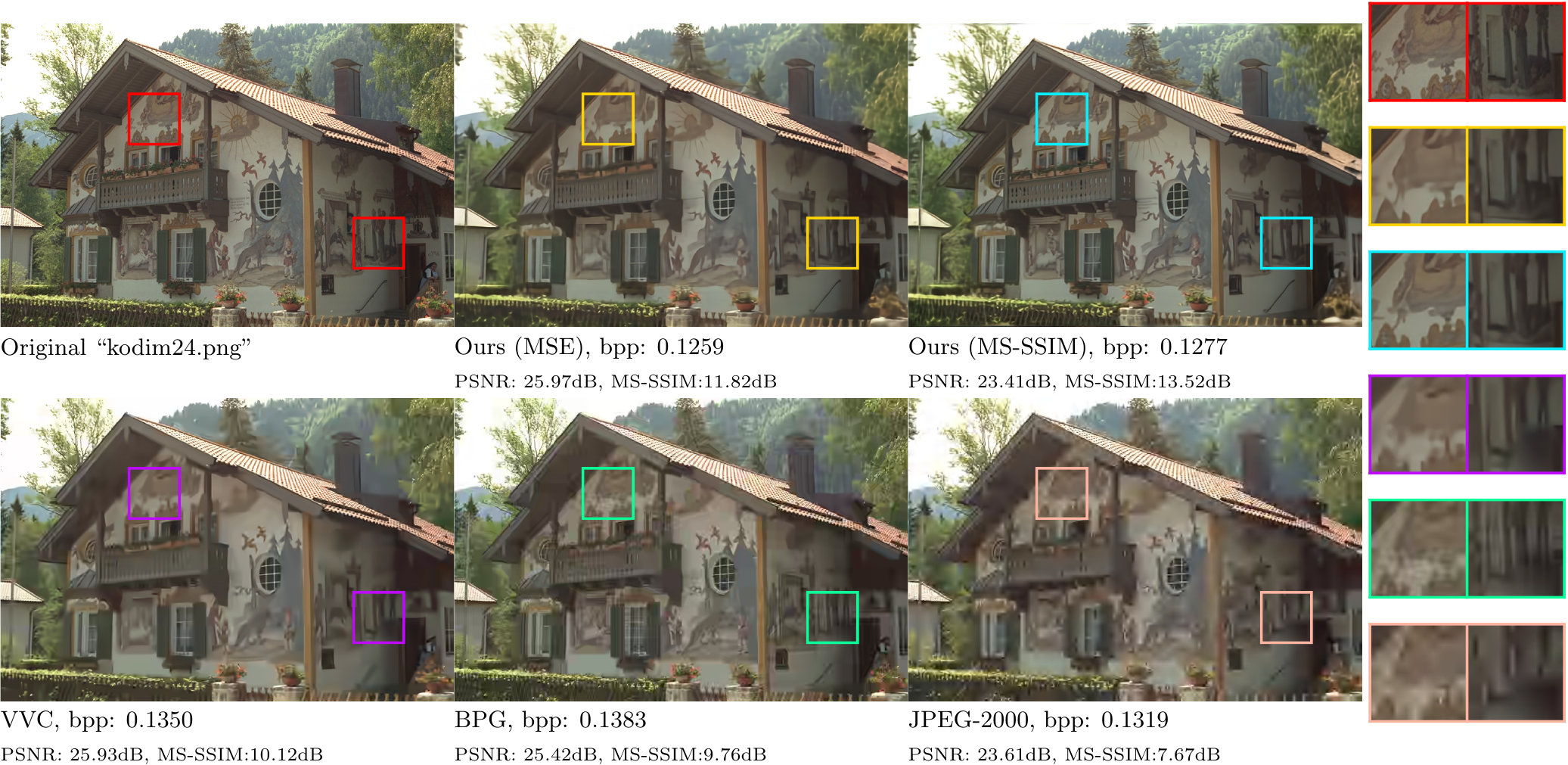}
  \caption{Visualization of \enquote{\texttt{kodim24.png}} for different codecs. Zoomed-in view on the right shows differences.}
  \label{Fig.Vis}
\end{figure*}

\subsection{Ablation Study}

To investigate impacts of proposed method, we conduct ablation study and report BD-rate \wrt original model (lower is better) and latency (\cref{Tab.AblationResult}):

\textbf{Impacts of cascaded estimation.} The level $L$ reflects how much parameters involved in estimation, \eg, $L = 1$ does not perform cascaded estimation (\enquote{$\mathit{\wo}$ cascaded}), while $L = 4$ will add an extra residual compared to original $L = 3$ model (\enquote{$4$-levels}). The first three rows of \cref{Tab.AblationResult} give results of three variants $L = 1,2,4$. As level increases, BD-rate continuously decreases. \enquote{$2$-levels} is much better than \enquote{$\mathit{\wo}$ cascaded} while \enquote{$4$-levels} obtains nearly no improvement compared to original model. The former indicates introducing cascaded estimation actually has a positive effect, while the latter tells us setting $L=3$ is sufficient otherwise model will be large and may hard to train. The \nth{2} column of table show latency between different models. Introducing more levels does not significantly slow model down, which indicates that cascaded estimation is not computational heavy for real scenario applications.

\textbf{Impacts of multi-codebook structure.} We use a global shared codebook as variant \enquote{$one$-codebook} to study impacts of multi-codebook structure. Results in the \nth{3} row of \cref{Tab.AblationResult} shows significant performance downgrade when using global codebook. It proves the effectiveness of multi-codebooks that model precise distributions since they adopt different parameters for different levels or groups.

\textbf{Impacts of quantization.} Quantization performance is affected in two way: a) Use a different similarity measure \eg cosine similarity (\enquote{$\mathit{cos}$-quantizer}) to define $\vect{\phi}$ in \cref{Eq.Phi}, b) Use a deterministic quantizer \ie same as \cite{VQGAN} or \cite{SoftQuantization} (\enquote{\cite{VQGAN}-quantizer, \cite{SoftQuantization}-quantizer}). The last three rows of \cref{Tab.AblationResult} shows difference of three quantizers. \enquote{$\mathit{cos}$-quantizer} adopts cosine similarity which is not a distance metric since it breaks the triangle inequality. We find this may cause performance drop. When training \enquote{\cite{VQGAN}-quantizer} or \enquote{\cite{SoftQuantization}-quantizer}, we find network is trapped in local-optima \ie most of vectors are quantized to a few codewords, and some codes are never assigned. We think this makes two kinds of variants have performance gap with ours.

\subsection{Visualization}
We pick image from Kodak to show compression quality. Compared codecs are JPEG-2000, BPG and VVC. All methods are set to $\mathit{bpp} \approx 0.13$ while compression ratio is about $185:1$. As \cref{Fig.Vis} shows, \enquote{\texttt{kodim24.png}} from Kodak dataset on the top-left is reference image. From zoomed-in view, we could find \enquote{Ours (MS-SSIM)} preserves more visual details, especially wall paintings and patterns. Meanwhile, it also achieves the highest MS-SSIM among all methods with the smallest $\mathit{bpp}$. Our MSE optimized model gives higher PSNR but is slightly blurred. It achieves comparable performance with VVC with a still small $\mathit{bpp}$. More perceptual measures and image comparisons are placed in supplementary materials.

We also give $2$-d projection visualization of $\vect{y}^1, \vect{y}^2$ on a toy model trained with $N=128, M=1, L=2, K=32$, shown in \cref{Fig.UMAPVisualization}. Specifically, latent vectors are extracted from $24$ Kodak images and projected to $2$-d points by UMAP~\cite{UMAP}. They are colored by codewords, \ie, two points are with same color if they are assigned to same codeword. The visualization satisfy our vectorized prior. Latents can be clustered by these codewords (left), while residuals are under similar distribution (right). Therefore, we can induce all latents to a unified, vectorized prior.

\section{Conclusion and Future Work}
In this paper, we propose a novel vectorized prior for variational image compression. We demonstrate latent vectors are correlated and can be induced to a unified multivariate Gaussian mixture. To perform estimation, proposed cascaded estimation with probabilistic vector quantization effectively approximate means and covariances. Furthermore, multi-codebooks are incorporated into above components to give an efficient compression procedure. Extensive experiments confirm effectiveness and efficiency of our proposed method. Future work will focus on variable-rate control with our proposed vectorized prior.

{\small
\paragraph{Limitation and Broader Impacts.} This work introduces a new perspective in neural image compression, which may inspire researchers to propose valuable future works. The high performance, low latency model may also benefit for real-life digital image storage or online multimedia contents. However, main limitations of our work are extra network parameters and computational resource requirement. Negative impacts involve vulnerability of model. We may give uncontrollable images under adversarial examples. Meanwhile, there seems to have no ethical issues or biases since the network is trained without supervision. However, training dataset does influence model with biased or sensitive images. Therefore, data should be checked to avoid potential issues.

\paragraph{Acknowledgements.} This work is supported by the National Natural Science Foundation of China (Grant No. 62020106008, No. 62122018, No. 61772116, No. 61872064), Sichuan Science and Technology Program (Grant No.2019JDTD0005).
}

\appendix

\section{Proof of \cref{Eq.MeanEstimation}}
For simplicity to prove \cref{Eq.MeanEstimation}, we will not consider residuals and cascaded estimation. Therefore, overall compression pipeline will become: $\vect{x} \xrightarrow{g_a} \vect{y} \xrightarrow{Q} \vect{\mathfrak{y}} = \tilde{\vect{y}} \xrightarrow{g_x} \tilde{\vect{x}}$, where $\vect{x}, \tilde{\vect{x}}$ are in two set $\vect{X}, \tilde{\vect{X}}$ with unknown distributions. According to rate-distortion theory, minimizing $\op{d}\left(\vect{x}, \tilde{\vect{x}}\right)$ is equivalent to maximizing mutual information between $\vect{X}$ and $\vect{\mathfrak{Y}}$:
\begin{equation}
\label{Eq.MutualInfoOri}
    \max{\mathcal{I}\left(\vect{X}; \; \vect{\mathfrak{Y}}\right)}.
\end{equation}
Since we do not add constraints on other network parameters, if network is fully trained, $\mathcal{I}\left(\vect{X}; \vect{Y}\right)$ should be maximized. Therefore \cref{Eq.MutualInfoOri} can be derived as:
\begin{equation}
\label{Eq.MutualInfo}
    \max{\mathcal{I}\left(\vect{Y}; \; \vect{\mathfrak{Y}}\right)}.
\end{equation}
Thus maximize the mutual information between latents and quantizeds. To solve $\vect{\mathfrak{Y}}$ or equivalence $\vect{\mathfrak{y}}$, \cref{Eq.MutualInfo} is wrote as a function \wrt $\vect{\mathfrak{y}}$:
\begin{equation}
\label{Eq.Integral}
    f\left(\vect{\mathfrak{y}}\right) = \int_{\vect{y}}{\sum_{\vect{\mathfrak{y}}}{p\left(\vect{y}, \vect{\mathfrak{y}}\right) \log{\frac{p\left(\vect{y}, \vect{\mathfrak{y}}\right)}{p\left(\vect{y}\right)p\left(\vect{\mathfrak{y}}\right)}}}}d\vect{y}.
\end{equation}
Noticed that $p\left(\vect{\mathfrak{y}}\right)$ is under Categorical prior whose all entries have the same probability \ie $p_{\vect{\mathfrak{Y}}}\left(\vect{\mathfrak{y}} = \vect{C}_i\right) = \nicefrac{1}{K}, \; 1 \leq i \leq K$~\cite{VQVAE}. \cref{Eq.Integral} can be simplified as:
\begin{equation}
\label{Eq.Simplify}
\begin{split}
    f\left(\vect{\mathfrak{y}}\right) &= \int_{\vect{y}}{\sum_{\vect{\mathfrak{y}}}{p\left(\vect{\mathfrak{y}}\right) p\left(\vect{y}\right) \log{\frac{p\left(\vect{\mathfrak{y}} \mid \vect{y}\right)}{p\left(\vect{\mathfrak{y}}\right)}}}}d\vect{y} \\
    &=\mathit{const} \cdot \int_{\vect{y}}{{p\left(\vect{y}\right) \log{p\left(\vect{\mathfrak{y}} \mid \vect{y}\right)}}}d\vect{y}.
\end{split}
\end{equation}
Since $p\left(\vect{y}\right)$ is not an variable of $f\left(\vect{\mathfrak{y}}\right)$, and $\log\left(\cdot\right)$ is monotonically increasing, maximizing \cref{Eq.Simplify} is equivalent to maximizing right part of function:
\begin{equation}
\label{Eq.Maximize}
    \max{f\left(\vect{\mathfrak{y}}\right)} \Leftrightarrow \max{\log{p\left(\vect{\mathfrak{y}} \mid \vect{y}\right)}}.
\end{equation}
This means when $\vect{y}$ is given to produce $\vect{\mathfrak{y}}$, that specific probability should be $\mathit{one}$ \ie fully confident. Recall that
\begin{equation}
\begin{split}
    p_{\vect{\mathfrak{Y}} \mid \vect{Y}}{\left( \vect{\mathfrak{y}} \mid \vect{y};\;\vect{C} \right)} &= \prod_{k=1}^{K}{\zeta{\left(\vect{\phi}\right)}_k^{\mathbbm{1}\left\{\vect{\mathfrak{y}} = \vect{C}_k\right\}}}, \\
    \mathit{where}\;\vect{\phi}_{k} &= -\norm{\vect{y} - \vect{C}_k}_2^2,\;1 \leq k \leq K.
\end{split}
\end{equation}
So,
\begin{equation}
\begin{split}
    &p_{\vect{\mathfrak{Y}} \mid \vect{Y}}{\left( \vect{\mathfrak{y}} = \vect{C}_k \mid \vect{y} \right)} \propto -\norm{\vect{y} - \vect{C}_k}_2^2, \\
    &p_{\vect{\mathfrak{Y}} \mid \vect{Y}}{\left( \vect{\mathfrak{y}} = \vect{C}_k \mid \vect{y} \right)} = 1 \Leftrightarrow \norm{\vect{y} - \vect{C}_k}_2^2 = 0.
\end{split}
\end{equation}
Therefore, if $\vect{y}$ has a nearest codeword $\vect{C}_k$, then $\vect{\phi}_k$ should be $\mathit{zero}$ in order to pick $\vect{C}_k$ with maximized confidence. So, for a subset $\vect{Y}_k = \set{\vect{y} \in \vect{Y} \mid \vect{\Phi}_k = 1}$, all $\vect{y}$s in this subset \enquote{\textit{pull}} codeword $\vect{C}_k$ to be close to them, making $\vect{C}_k$ to be the mean embedding of $\vect{Y}_k$. So \cref{Eq.MeanEstimation} is derived.

\begin{figure*}[t]
\centering
\includegraphics[height=0.7\columnwidth]{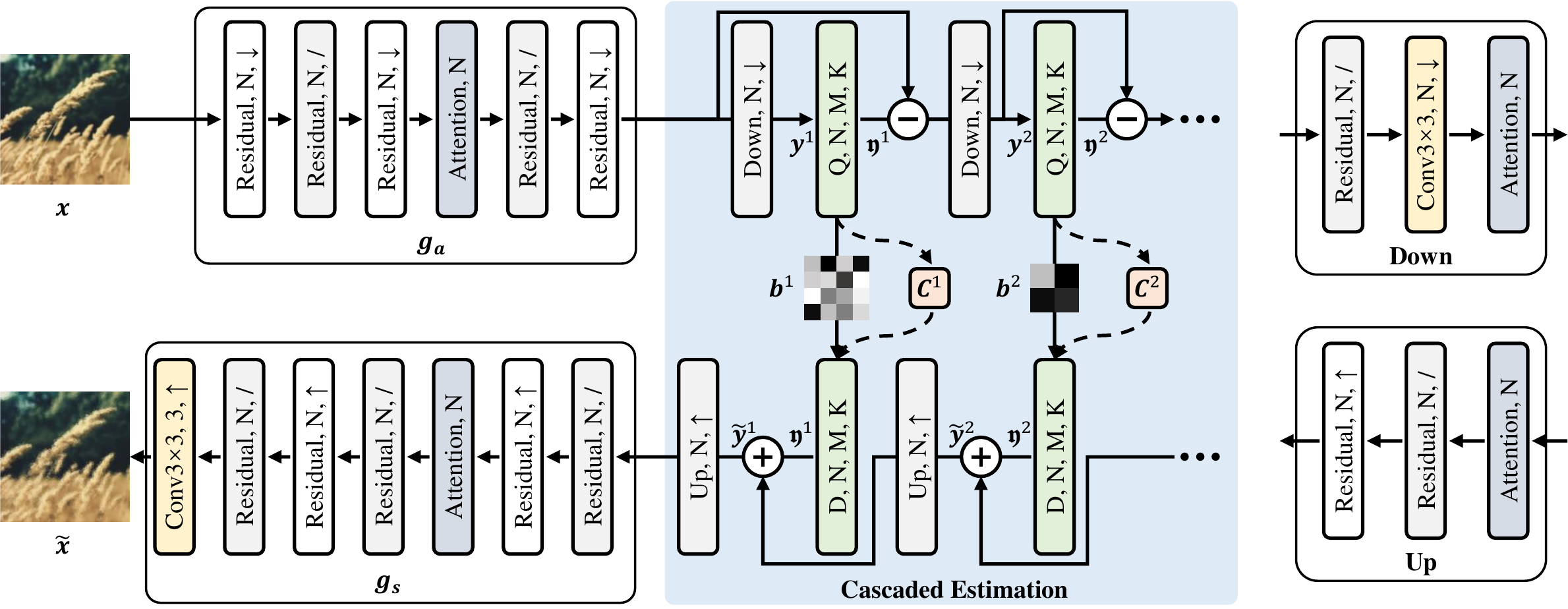}
\caption{The detailed framework. \enquote{Residual}, \enquote{Attention} are residual block and attention block from \cite{GMMAttention}. \enquote{Down} and \enquote{Up} blocks are placed on the right. \enquote{N} is number of channels. \enquote{$\downarrow$} means output is $2\times$ down-sampled and vice-versa. \enquote{M}, \enquote{K} are codebook sizes ($M$ sub-codebooks, $K$ codewords for each.).}
\label{Fig.DetailedFramework}
\end{figure*}

\begin{figure*}[t]
    \centering
    \includegraphics[height=0.7\columnwidth]{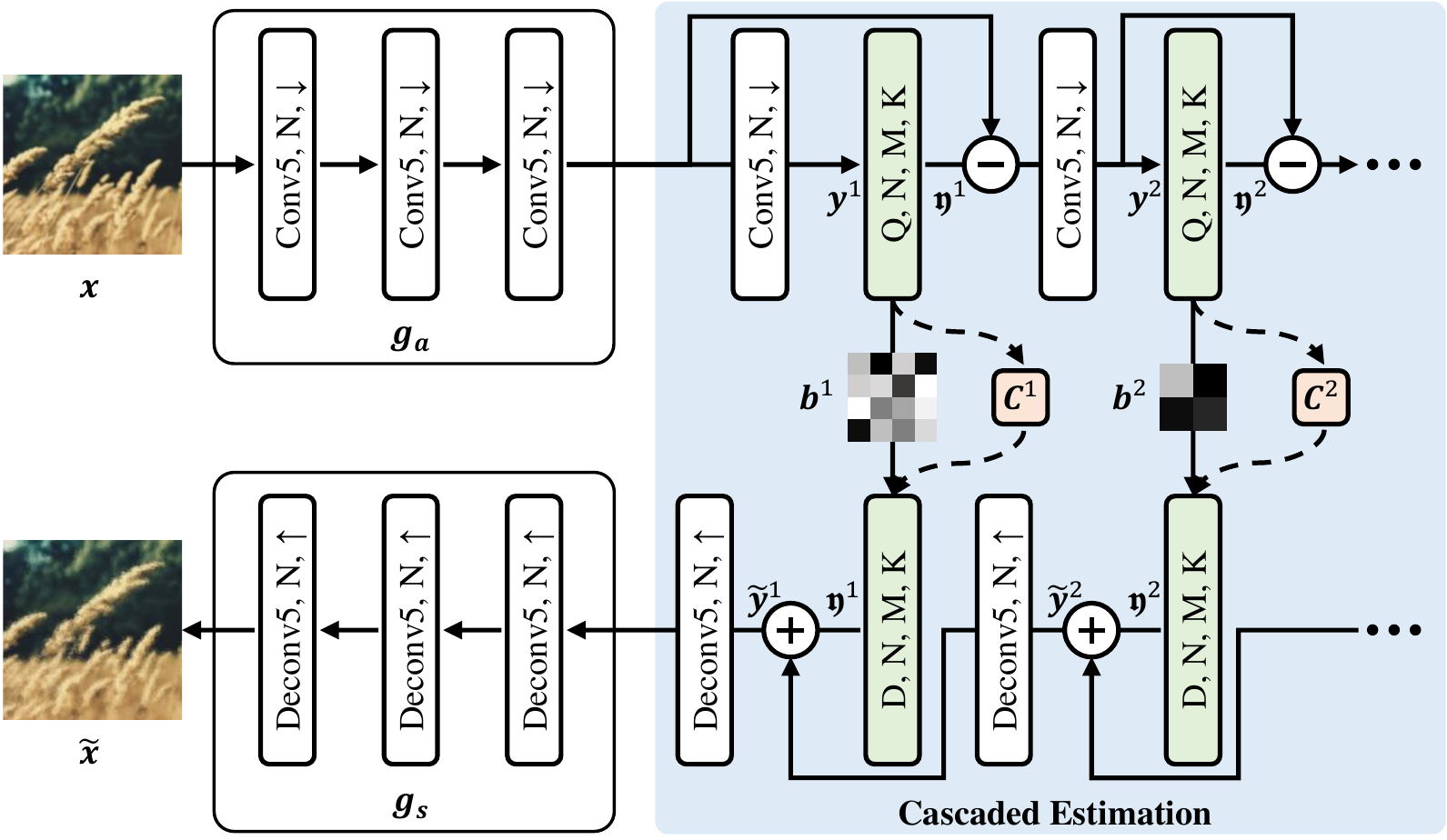}
    \caption{An additional model with $\mathit{Conv5}$ and $\mathit{Deonv5}$ layers. The structure is similar as Ball{\'e}'s~\cite{ScaleHyper} and Minnen's~\cite{JointHyper} in order to test generalization ability of our method. Specifically, $\mathit{Conv5}$ uses $\mathit{kernel\_size} = 5\times5, \mathit{stride} = 2, \mathit{padding} = 2$, and output size is $2\times$ smaller than input. $\mathit{Deconv5}$ is reverse of $\mathit{Conv5}$.}
    \label{Fig.Conv5Framework}
\end{figure*}

\begin{table}[t]
\centering
\resizebox{\columnwidth}{!}{
\begin{tabular}{@{}cc|rrrr@{}}
\toprule
\multicolumn{2}{c|}{\multirow{3}{*}{Methods}}              & \multicolumn{4}{c}{Latency ($\mathit{ms}$)}               \\
\multicolumn{2}{c|}{}                                      & \multicolumn{2}{c}{Encoder} & \multicolumn{2}{c}{Decoder} \\ \cmidrule(lr){3-4}\cmidrule(lr){5-6}
\multicolumn{2}{c|}{}                                      &\multicolumn{1}{c}{Abs}&\multicolumn{1}{c}{Rel}&\multicolumn{1}{c}{Abs}&\multicolumn{1}{c}{Rel}\\ \midrule\midrule
\multicolumn{2}{c|}{Ball{\'e}'18}                   & $30.66$  & $1.09\times$& $35.54$  & $1.21\times$ \\\midrule
\multirow{3}{*}{Minnen'18}&$\wo$                    & $32.89$  &$1.17\times$ & $36.24$  &$1.24\times$ \\
                               &$\rightarrow$       & $2656.66$ &$94.58\times$& $1799.47$&$61.36\times$ \\
                             &$\vcdice{5}$          & $59.13$  &$2.11\times$ & $40.40$  &$1.38\times$  \\\midrule
\multirow{2}{*}{Cheng'20}&$\rightarrow$             & $2697.58$&$96.04\times$& $1835.80$&$62.60\times$ \\
                           &$\vcdice{5}$            & $94.11$  &$3.35\times$ & $88.04$  &$3.00\times$  \\\midrule
\multicolumn{2}{c|}{Ours}                           & $\mathbf{28.09}$  &$\mathbf{1.00\times}$ & $\mathbf{29.32}$  &$\mathbf{1.00\times}$  \\
\multicolumn{2}{c|}{Our Additional}                 & $\mathbf{12.03}$  &$\mathbf{0.43\times}$ & $\mathbf{13.37}$  &$\mathbf{0.46\times}$  \\\bottomrule
\end{tabular}
}
\caption{Encoding and decoding latency comparisons for image size $768 \times 512$. Our additional model achieves even faster speed than our main model.}
\label{Tab.LatencyConv5}
\end{table}
\begin{table}[!t]
\centering
\resizebox{\columnwidth}{!}{
\begin{tabular}{@{}r|cccccccc@{}}
\toprule
\multicolumn{9}{c}{Variants}                         \\ \midrule\midrule
$M$ & $1$  & $2$   & $4$   & $6$   & $8$    & $12$   & $16$   & $24$   \\
$K$ & $64$ & $128$ & $256$ & $512$ & $1024$ & $2048$ & $4096$ & $8192$ \\ \bottomrule
\end{tabular}
}
\caption{Models with different codebook sizes. We test codebooks from small to large to validate scalability.}
\label{Tab.Size}
\end{table}

\begin{figure*}[!htp]
\centering
\begin{code}
class Quantizer(Module):
    """
    Quantizer with `m` sub-codebooks,
        `k` codewords for each, and
        `n` total channels.
    Args:
        m (int): Number of sub-codebooks.
        k (int): Number of codewords for each sub-codebook.
        n (int): Number of channels of latent variables.
    """
    def __init__(self, m: int, k: int, n: int):
        super().__init__()
        # A codebook, feature dim `d = n // m`.
        self._codebook = Parameter(torch.empty(m, k, n // m))
        self._initParameters()

    def forward(self, x: Tensor, t: float = 1.0) -> (Tensor, Tensor):
        """
        Module forward.
        Args:
            x (Tensor): Latent variable with shape [b, n, h, w].
            t (float, 1.0): Temperature for Gumbel softmax.
        Return:
            Tensor: Quantized latent with shape [b, n, h, w].
            Tensor: Binary codes with shape [b, m, h, w].
        """
        b, _, h, w = x.shape
        # [b, m, d, h, w]
        x = x.reshape(b, len(self._codebook), -1, h, w)
        # [b, m, 1, h, w], square of x
        x2 = (x ** 2).sum(2, keepdim=True) |\label{line.31}|
        # [m, k, 1, 1], square of codebook
        c2 = (self._codebook ** 2).sum(-1, keepdim=True)[..., None]
        # [b, m, d, h, w] * [m, k, d] -sum-> [b, m, k, h, w]
        # dot product between x and codebook
        inter = torch.einsum("bmdhw,mkd->bmkhw", x, self._codebook)
        # [b, m, k, h, w], pairwise L2-distance
        distance = x2 + c2 - 2 * inter |\label{line.37}|
        # [b, m, k, h, w], distance as logits to sample
        sample = F.gumbel_softmax(-distance, t, hard=True, dim=2) |\label{line.39}|
        # [b, m, d, h, w], use sample to find codewords
        quantized = torch.einsum("bmkhw,mkd->bmdhw", sample, self._codebook)
        # back to [b, n, h, w]
        quantized = quantized.reshape(b, -1, h, w)
        # [b, n, h, w], [b, m, h, w], quantizeds and binaries
        return quantized, sample.argmax(2) |\label{line.45}|
\end{code}
\caption{Minimal implementation of our probabilistic vector quantization.}
\label{Fig.Code}
\end{figure*}

\begin{figure*}[t]
    \centering
    \includegraphics[width=0.8\textwidth]{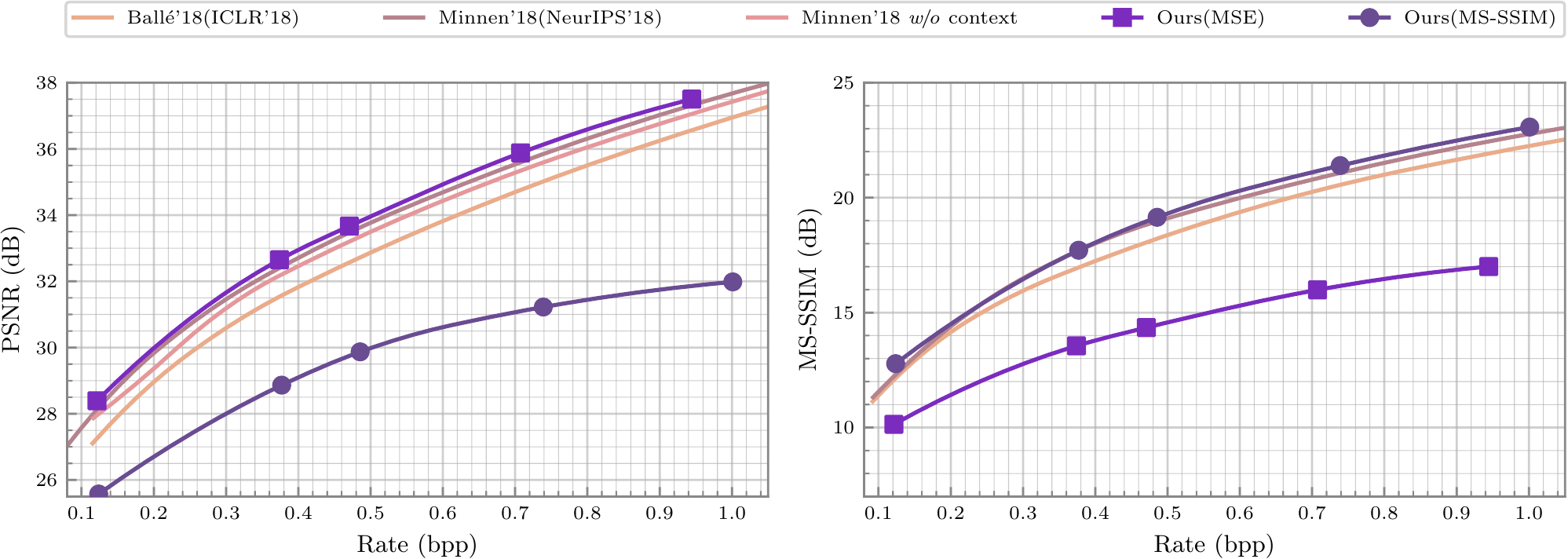}
    \caption{Rate-Distortion performance with our additional $\mathit{conv5}$-based model on Kodak dataset. Ours is slightly better than Ball{\'e}'18 and Minnen'18, which have similar backbone.}
    \label{Fig.RDConv5}
\end{figure*}

\begin{figure}[t]
    \centering
    \includegraphics[width=1\columnwidth]{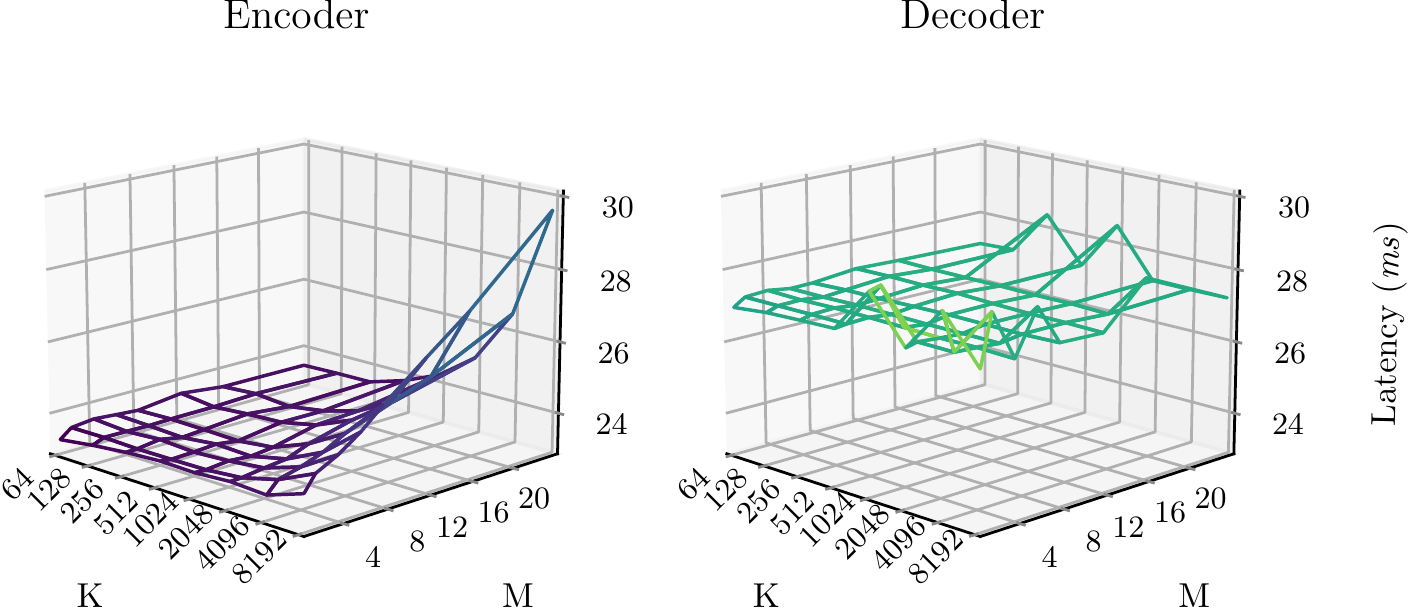}
    \caption{Latencies \wrt codebook sizes. Some bumps in decoder is considered to be within margin of error.}
    \label{Fig.Latency}
\end{figure}
\begin{figure}
    \centering
    \includegraphics[width=\columnwidth]{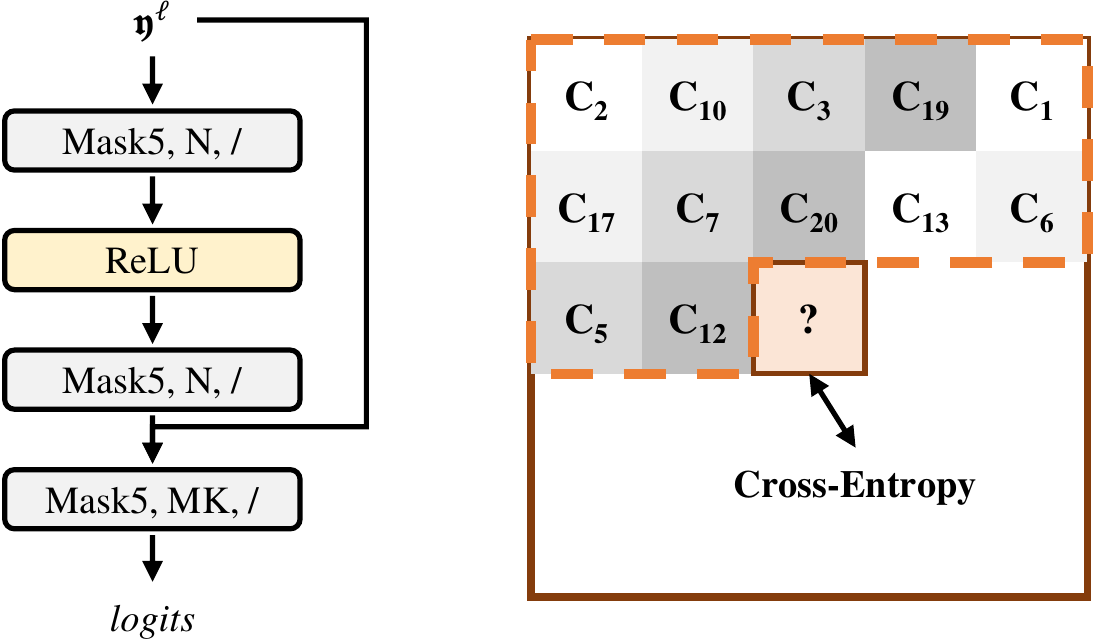}
    \caption{\textbf{Left}: Residual-based masked convolutional block. We use \texttt{MaskedConv} with kernel size $5\times 5$. The last layer produces $M\times K$ logits and trained by picked codeword indices and cross-entropy loss. \textbf{Right}: Demonstration of masked prediction. It uses left and top $\vect{\mathfrak{y}}$ as context information to predict index of next picked codeword.}
    \label{Fig.MaskedConv}
\end{figure}

\section{Model Architecture}
Now, we show our detailed model design in \cref{Fig.DetailedFramework}. We adopt similar structure from \cite{GMMAttention} \ie Residual blocks and attention blocks. You could check structures in their paper. \enquote{Down} and \enquote{Up} blocks are placed on the right. \enquote{$\downarrow$} means output is $2\times$ down-sampled and vice-versa. $N$ the number of channels, $M$, $K$ the codebook sizes ($M$ sub-codebooks, $K$ codewords for each.).

For instance, given an image of size $768 \times 512$, we first $16\times$ down-sample it to $96 \times 64$, then go to cascaded estimation. Each $\vect{y}$ is obtained by one more down-sampling. So, size of $\vect{y}^1$ is $48 \times 32$, $\vect{y}^2$ is $24 \times 16$, \etc, and vice versa for decoding.

\paragraph{How to Calculate $\sup\mathit{bpp}$?} As shown above, output code size will be $16^2\times, 32^2\times, 64^2\times$ smaller than original images. According to Sec.4, the upper bound of $\mathit{bpp}$ is:
\begin{equation}
    M \cdot \frac{\sum_{l}{\log_{2}{K} \cdot h^\ell \cdot w^\ell}}{H \cdot W}.
\end{equation}
For example, when we employ model \textnumero1, where $M = 2$ and $K = \left[8192, 2048, 512\right]$, the above result is:
\begin{equation}
     2 \cdot \left(\frac{13}{16^2} + \frac{11}{32^2} + \frac{9}{64^2}\right) \approx 0.1274
\end{equation}

\section{Implementation}
To implement a probabilistic vector quantization with multi-codebooks, we could seek help from a few PyTorch built-in functions such as \texttt{einsum} and \texttt{gumbel\_softmax}. Our implementation is shown in \cref{Fig.Code}. Specifically, to calculate the pair-wise Euclidean distance in order to produce $\vect{\phi}$, we can use the expanded version for speed up, \eg, for two matrix $\vect{U} \subseteq \mathbb{R}^{k_1 \times n}, \vect{V} \subseteq \mathbb{R}^{k_2 \times n}$, the pairwise distance $\vect{D} \subseteq \mathbb{R}^{k_1 \times k_2}$ is calculated by: $\vect{U}^2 + \vect{V}^2 - 2\vect{U}\vect{V}^\intercal$. We could utilize \texttt{einsum} to perform above calculation in $M$ ways separately with very few line of codes (line \textnumero~\texttt{\ref{line.31} $\sim$ \ref{line.37}}). Then, the calculated distance will be input of \texttt{gumbel\_softmax} to sample one-hot vectors (line \textnumero~\texttt{\ref{line.39}}). The indices of where \enquote{$one$}s present are collected as $\vect{b}$ (line \textnumero~\texttt{\ref{line.45}}).

\section{Additional Experiments}
We also conduct a few additional experiments to investigate detailed latencies, model generalization ability, \etc.
\subsection{Compression Latencies \wrt Codebook Size}
We conduct latency tests of our models by varying $M, K$ in codebook, to see if codebook size could affect model efficiency. Specifically, we set $N = 192$, $L = 1$, $M$ varies from $1$ to $24$ and $K$ varies from $64$ to $8192$. Detailed settings are placed in \cref{Tab.Size} while results are placed in \cref{Fig.Latency}.

From \cref{Fig.Latency}, we could draw following conclusions. Firstly, encoder's latency is linearly correlated to $K$. This is because computation of $\vect{\phi}$ consumes $\mathcal{O}\left(NKD\right)$ time complexity.

Decoder's latency is smooth and flat. Since it is not affected by $K$ or $M$. Decoding only involves $\mathcal{O}\left(1\right)$ lookup and operations between sub-codebooks are highly paralleled. Therefore, no matter how many codewords are employed in quantization, decoding can be still treated as $\mathcal{O}\left(1\right)$ roughly.


\subsection{Additional Perceptual Evaluations}
To make a comprehensive study on image restoration quality of out network as well as other codecs, we notice that there are a lot of perceptual metrics can be adopted. In this study, we choose $\Delta E$~\cite{deltaE}, LPIPS~\cite{lpips} and Inception score (IS)~\cite{inceptionscore}\footnote{We use their open-source PyTorch implementations.}. We pick \enquote{VVC}, \enquote{Cheng'20} and \enquote{ours} to test since they have similar performance in main paper. Due to limitation of computation resources, we only test with our model \textnumero2 and tune quantization parameter of other codecs to target similar $\mathit{bpp}$. Results are shown in \cref{Tab.AdditionalMetric}. From this table, we could confirm codecs that target MSE generally perform worse than MS-SSIM on LPIPS and IS scores. In contrast, MSE models have lower $\Delta E$ than MS-SSIM models.

\begin{table}[t]
\centering
\resizebox{\columnwidth}{!}{
\begin{tabular}{@{}l|cccc@{}}
\toprule
Codecs             & BPP & $\Delta E$ & LPIPS  & IS \\ \midrule\midrule
VVC VTM 14.2       & $0.1455$ &$\textbf{3.927}$& $0.159$ & $3.593$ \\
Cheng'20 (MSE)     & $0.1281$ & $4.577$    & $0.161$ & $3.542$ \\
Cheng'20 (MS-SSIM) & $0.1279$ & $4.782$    & $0.149$ & $3.791$ \\
Ours (MSE)         & $0.1234$ & $4.505$    & $0.163$ & $3.535$ \\
Ours (MS-SSIM)     & $0.1256$ & $4.966$    & $\textbf{0.144}$ & $\textbf{3.875}$ \\ \bottomrule
\end{tabular}
}
\caption{Perceptual comparisons between VVC, Cheng'20 and ours on Kodak dataset. The results indicate that deep models targeting MS-SSIM generally perform better than MSE on LPIPS and IS. And in contrast, MSE models as well as VVC perform better on $\Delta E$.}
\label{Tab.AdditionalMetric}
\end{table}

\subsection{R-D Performance with Other Backbones}

In main paper, we only report R-D performance based on \cite{GMMAttention}'s backbone. It is important to test with other backbones to evaluate the generalization ability of our method. Therefore, we design an additional model, as shown in \cref{Fig.Conv5Framework}. Specifically, we use $\mathit{Conv5}$ and $\mathit{Deconv5}$ with $5\times5$ kernels to perform $2\times$ down-sampling and up-sampling.

Results on Kodak dataset are placed in \cref{Fig.RDConv5}. We only compare our additional model with methods that have similar backbone \ie Ball{\'e}'18~\cite{ScaleHyper} and Minnen'18~\cite{JointHyper}. Similar as results in main paper, our method has a slightly better R-D performance against Ball{\'e}'18 and Minnen'18. These results indicate our method is suitable for different backbones. It is foreseeable that our method would be effective if incorporates with other backbones \eg \cite{ICCV21}.

We also test latencies with this additional model, placed in \cref{Tab.LatencyConv5} at the last row. From the table we find that latencies of the additional model are further reduced. This is because our additional model has fewer layers than main model. Compared to Ball{\'e}'18 and Minnen'18, our model is much faster. We will release two types of models in the future.

\begin{table}[t]
\centering
\begin{tabular}{@{}ccc@{}}
\toprule
Context $\mathit{Acc.}$ & $\mathit{bpp\;w/}$ & $\mathit{bpp}\;\wo$ \\ \midrule\midrule
$2.24\%$                & $0.1262$           & $0.1265$  \\ \bottomrule
\end{tabular}
\caption{Prediction accuracy of the auxiliary context model. With context model, we only obtain $2.24\%$ accuracy for context prediction on the average among all levels and all groups, and $\mathit{bpp}$ is nearly the same with original model.}
\label{Tab.Predict}
\end{table}

\subsection{Incorporating with Auxiliary Context Model}
In main paper, we claim that our model does not need auxiliary context models for side information prediction. But we still want to know whether context models could help for better Rate-Distortion performance. To incorporate with a context model, we adopt the widely-used PixelCNN~\cite{PixelCNN}. Specifically, we build a residual-like block with full of $\mathit{MaskedConv}$ layers (\cref{Fig.MaskedConv}) as the causal prediction network. We insert these blocks directly after $\vect{\mathfrak{y}}^\ell$ on every level. Then, they produce $M \times K$ logits and are trained with picked codeword indices and cross-entropy loss. This procedure is also demonstrated in \cref{Fig.MaskedConv}.

To evaluate how well these introduced networks predict, prediction accuracy of next picked codeword is calculated. For instance, if they could predict $50\%$ of picked codewords' indices, $\mathit{bpp}$ will be reduced $50\%$ approximately.

Results are reported in \cref{Tab.Predict}. With context model, we only obtain $2.24\%$ accuracy for context prediction on the average among all levels and all groups, and $\mathit{bpp}$ is nearly the same with original model. This indicates that our model has encoded binary codes with high information entropy. Introducing extra context model do not further reduce rate.

\begin{figure}[t]
  \centering
  \begin{subfigure}{0.494\columnwidth}
    \includegraphics[width=\linewidth]{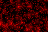}
    \caption{Perturbed area. $15\%$ codes are changed to random new values.}
    \label{Fig.PerturbArea}
  \end{subfigure}
  \hfill
  \begin{subfigure}{0.494\columnwidth}
    \includegraphics[width=\linewidth]{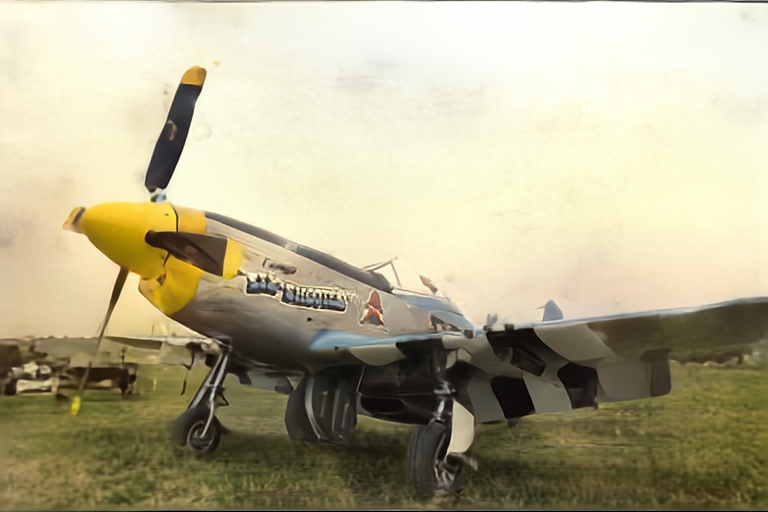}
    \caption{Visualization result after perturbation. BPP: $0.1278 \rightarrow 0.1283$.}
    \label{Fig.PerturbResult}
  \end{subfigure}
  \caption{Effects of code perturbation.}
  \label{Fig.Perturb}
\end{figure}
\begin{figure*}[t]
  \centering
  \begin{subfigure}{0.24\linewidth}
    \includegraphics[width=\linewidth]{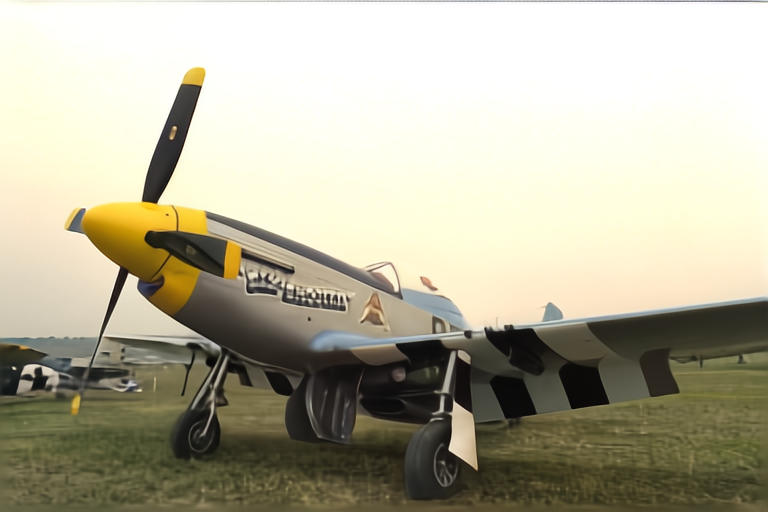}
    \caption{Clean image. BPP: $0.1280$}
  \end{subfigure}
  \begin{subfigure}{0.24\linewidth}
    \includegraphics[width=\linewidth]{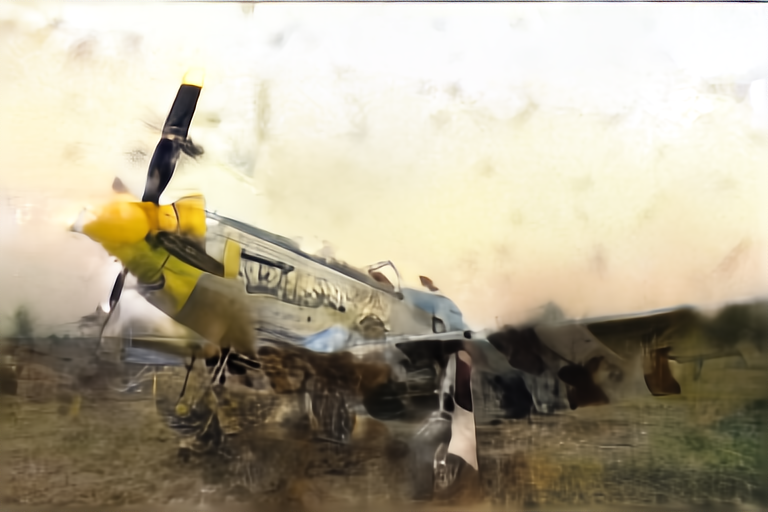}
    \caption{$25\%$ perturbed. BPP: $0.1285$}
  \end{subfigure}
  \begin{subfigure}{0.24\linewidth}
    \includegraphics[width=\linewidth]{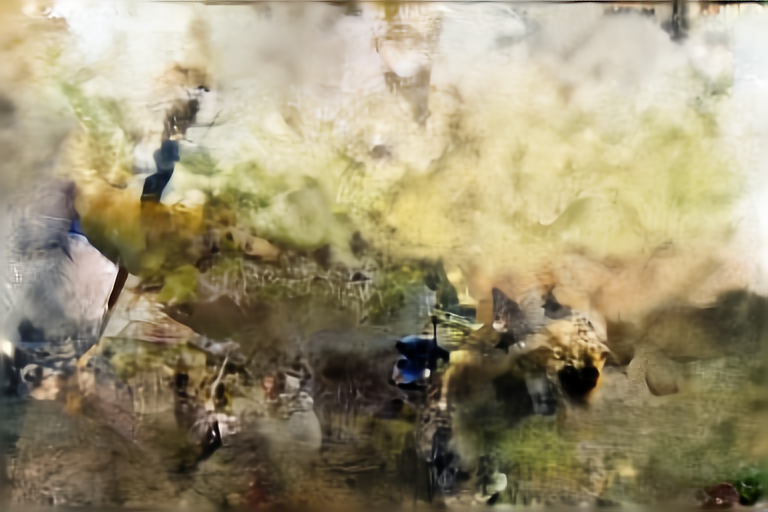}
    \caption{$50\%$ perturbed. BPP: $0.1288$}
  \end{subfigure}
  \begin{subfigure}{0.24\linewidth}
    \includegraphics[width=\linewidth]{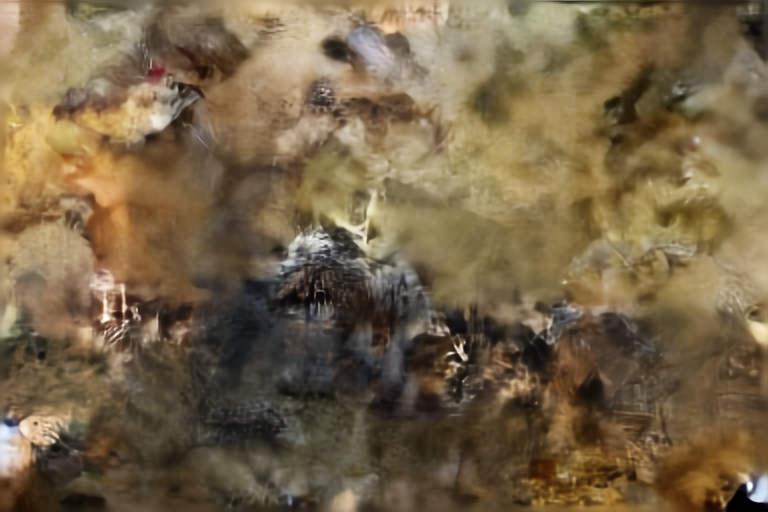}
    \caption{$75\%$ perturbed. BPP: $0.1294$}
  \end{subfigure}
  \caption{Restored images with various perturbation proportions.}
  \label{Fig.NoiseAll}
\end{figure*}

\subsection{Effects of Code Perturbation}

As mentioned in limitations and broader impacts, we could craft images that corrupt vectorized prior by \eg adversarial attack. Therefore, a simple study is conducted by perturbing partial of compression codes to simulate this approach.

Specifically, We randomly perturb $15\%$ of $\vect{b}$ that produced by our MS-SSIM model \textnumero$1$. Result based on \cref{Fig.Vis} is shown in \cref{Fig.Perturb}. In \cref{Fig.PerturbArea}, lighter area indicates more codes are perturbed. \cref{Fig.PerturbResult} is reconstruction result that appears to be artifacts on it. For whole Kodak dataset, after perturbation, $\mathit{bpp}$ increases $0.1256\!\rightarrow\!0.1267$ and MS-SSIM decreases $14.33\!\rightarrow\!8.83$.

From above observations, firstly we think these reveal model's ability to choose appropriate codes for good rate-distortion. If any codes are misplaced, not only performance will drop, but also rate will increase. And intuitively, if perturbation rate increases, model performance will continuously drop. This can be confirmed by following experiments in \cref{Fig.NoiseAll}. We think this is a valuable research problem and would like to conduct future studies on robustness and mechanisms of proposed vectorized prior.

\subsection{Visualization}

\begin{figure*}[t]
\centering
\includegraphics[width=0.7\linewidth]{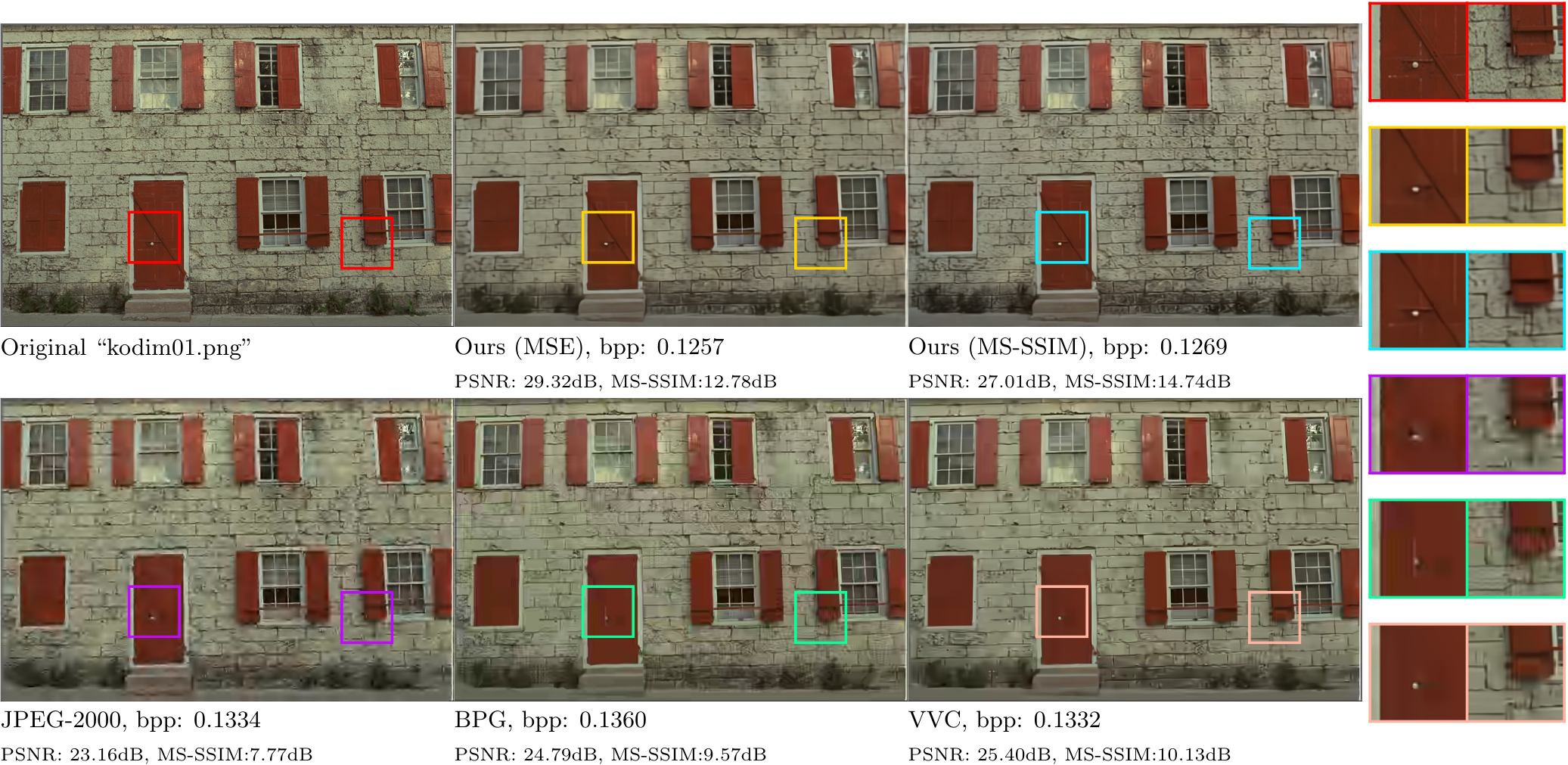}
\caption{Comparisons of \enquote{\texttt{kodim01.png}} with other codecs.}
\label{Fig.Vis1}
\end{figure*}

\begin{figure*}[t]
\centering
\includegraphics[width=0.7\linewidth]{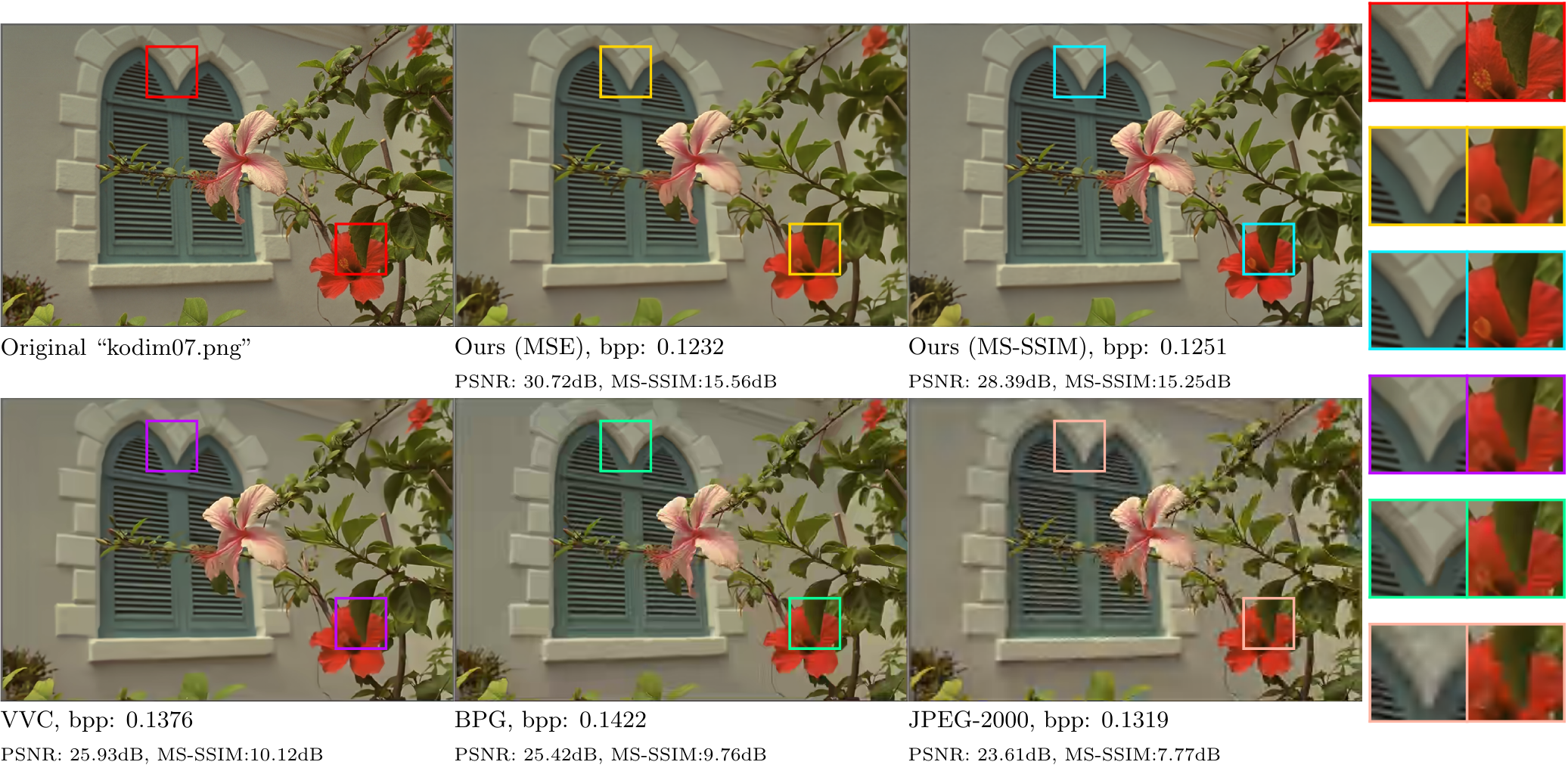}
\caption{Comparisons of \enquote{\texttt{kodim07.png}} with other codecs.}
\label{Fig.Vis2}
\end{figure*}

\begin{figure*}[t]
\centering
\includegraphics[width=0.5\linewidth]{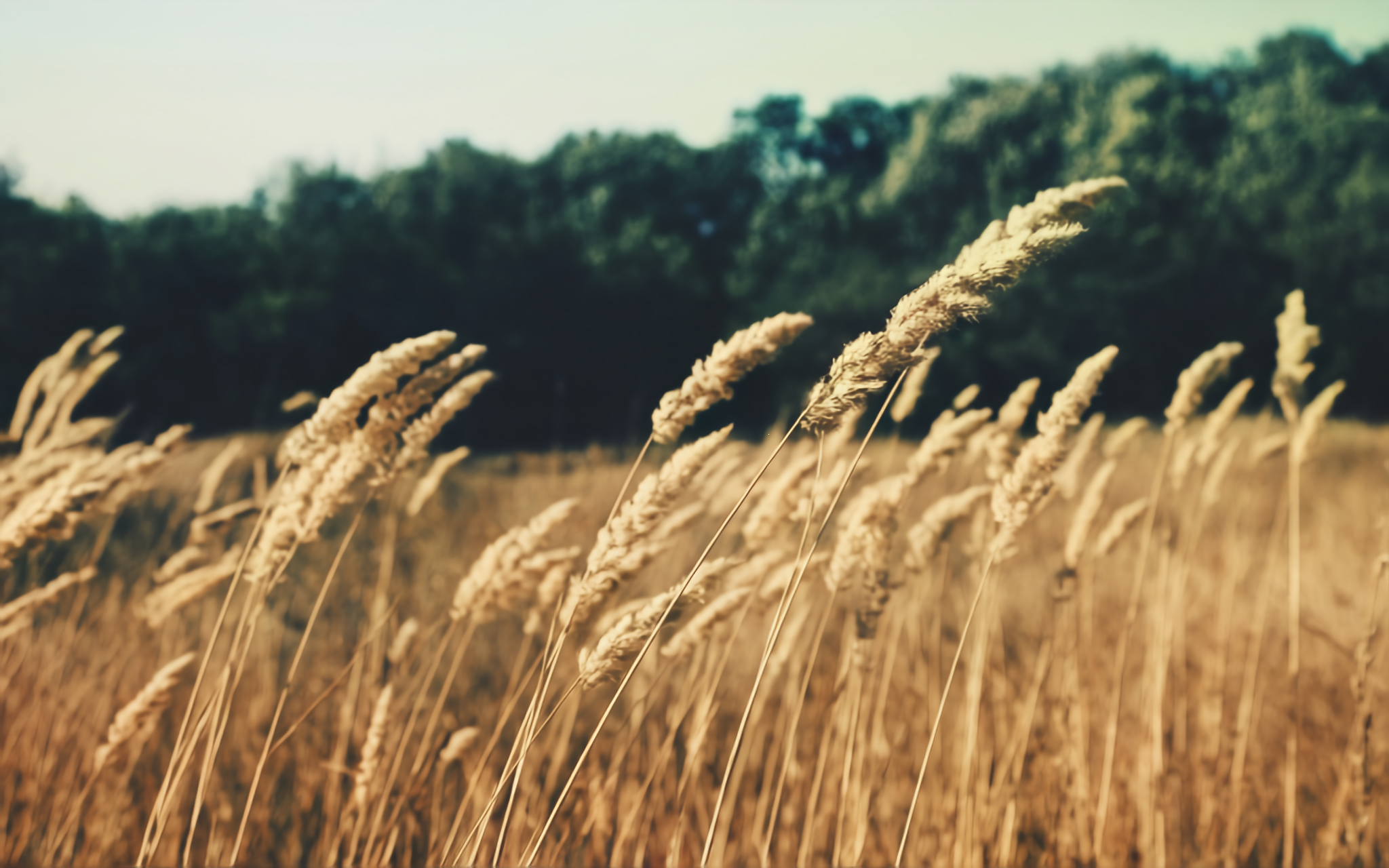}
\caption{Ours $\text{bpp} = 0.1236$, $\text{PSNR} = 34.29\text{dB}$, $\text{MS-SSIM} = 19.40 \text{dB}$.}
\label{Fig.Vis3}
\end{figure*}

\begin{figure*}[t]
\centering
\includegraphics[width=0.5\linewidth]{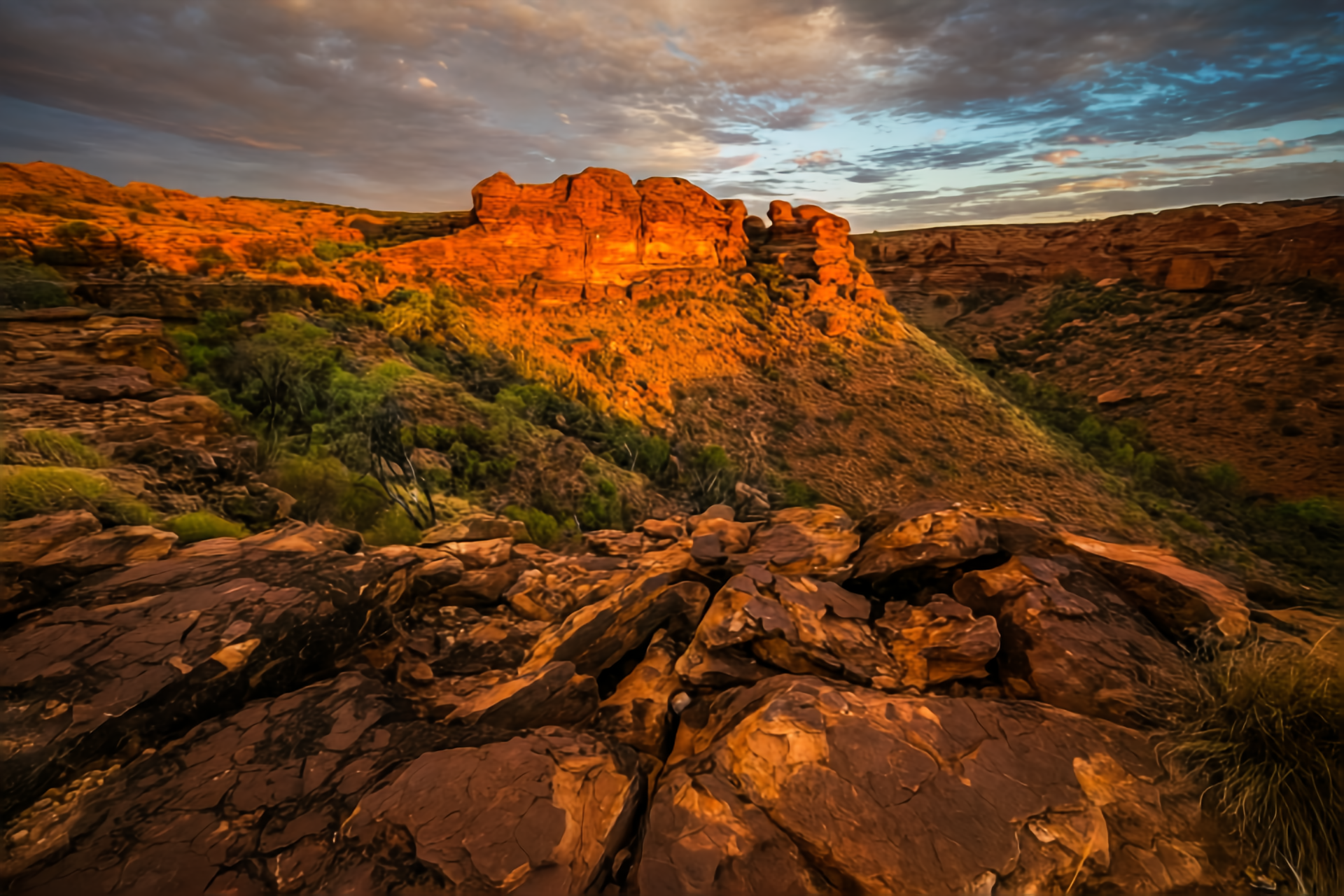}
\caption{Ours $\text{bpp} = 0.1265$, $\text{PSNR} = 26.95\text{dB}$, $\text{MS-SSIM} = 11.64 \text{dB}$.}
\label{Fig.Vis4}
\end{figure*}

\begin{figure*}[t]
\centering
\includegraphics[width=0.5\linewidth]{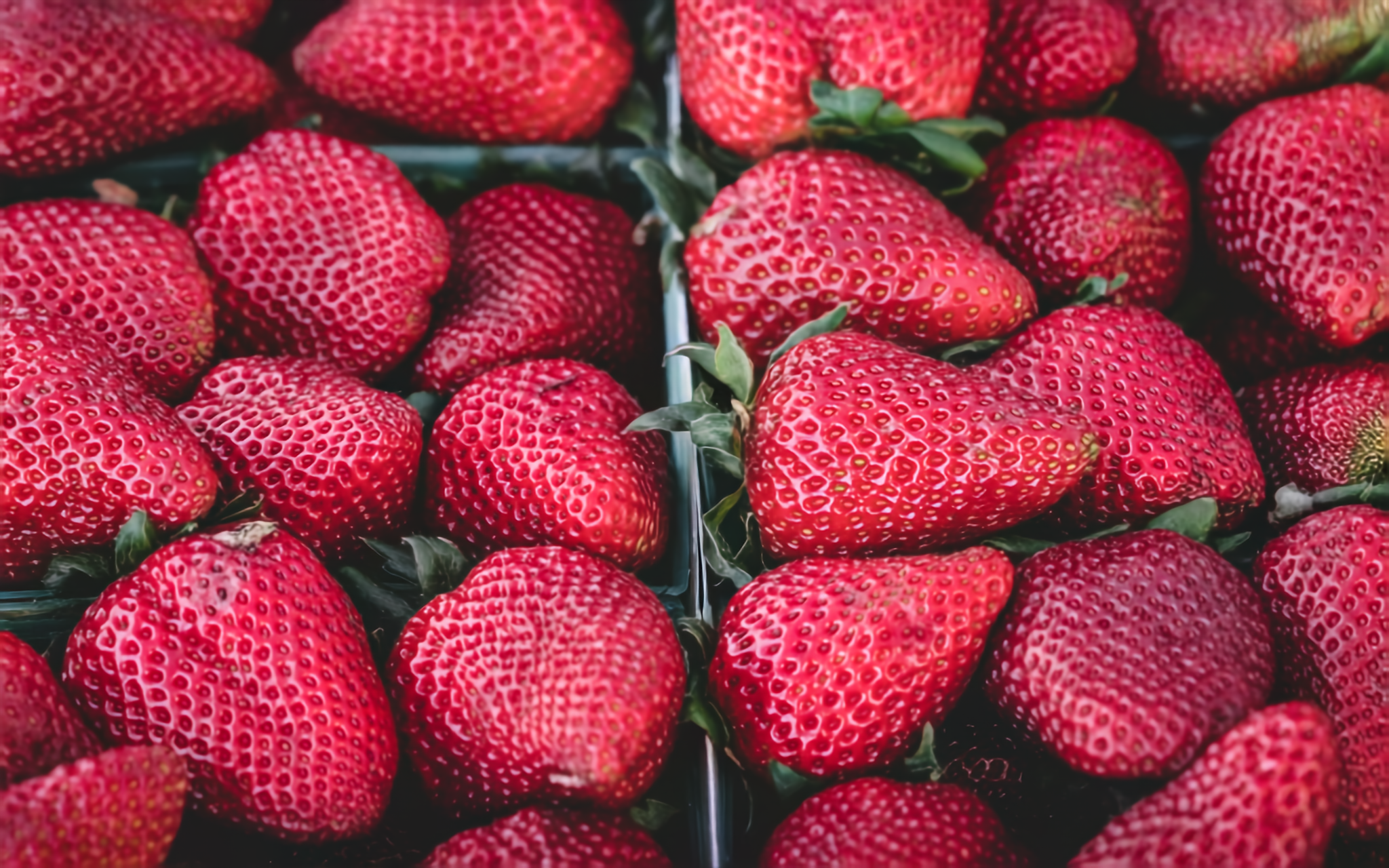}
\caption{Ours $\text{bpp} = 0.1259$, $\text{PSNR} = 27.04\text{dB}$, $\text{MS-SSIM} = 13.51 \text{dB}$.}
\label{Fig.Vis5}
\end{figure*}

We further pick $2$ images in Kodak and $3$ images in CLIC Professional valid set for comprehensive visualization in \crefrange{Fig.Vis1}{Fig.Vis5}. From these figures we could find our model preserves rich details.


\clearpage
\clearpage
\clearpage
\clearpage
\clearpage
\clearpage
\clearpage
\clearpage
\clearpage
\clearpage
\clearpage
\clearpage
\clearpage
\clearpage
\clearpage
\clearpage
\clearpage
\clearpage
\clearpage
\clearpage
\newpage
{\small
\bibliographystyle{ieee_fullname}
\bibliography{egbib}
}

\end{document}